%% file: main.tex
\documentclass[letterpaper, 10 pt, journal, twoside]{IEEEtran} 
\usepackage[utf8]{inputenc}
\IEEEoverridecommandlockouts                              

\usepackage{mathtools}
\usepackage{graphicx}
\usepackage{booktabs}
\usepackage{amsfonts}
\usepackage{subfig}
\usepackage[linesnumbered,ruled,vlined]{algorithm2e}
\usepackage{multirow}
\usepackage{color}
\usepackage[bookmarks=false,hyperfootnotes=false]{hyperref}
\usepackage{setspace} 
\usepackage{xcolor}
\usepackage{balance}

\usepackage{amssymb}

\usepackage[font={small}]{caption}

\SetKwInput{KwData}{Global params.}
\SetKw{return}{return}
\SetKw{Break}{break}

\SetCommentSty{mycommfont}

\title{Asymptotically optimal path planning with an approximation of the omniscient set}

\author{Jon\'{a}\v{s} K\v{r}\'\i\v{z}, Vojt\v ech Von\' asek
\thanks{Manuscript received: September, 16, 2024; Revised January, 2, 2025; Accepted January, 31, 2025.}
\thanks{This paper was recommended for publication by Editor Aniket Bera upon evaluation of the Associate Editor and Reviewers' comments.
This work was supported by the Czech Science Foundation (GA{\v C}R) under project No. 24-12360S, by the European Union under the project Robotics and advanced industrial production (no. CZ.02.01.01/00/22\_008/0004590), and by CTU grant no. SGS23/177/OHK3/3T/13.
Computational resources were provided by the e-INFRA
CZ project (ID:90254), supported by the Ministry of Education, Youth and Sports of the Czech Republic.
}
\thanks{
The authors are with the Faculty of Electrical
Engineering, Czech Technical University in Prague, Czech Republic. }
\thanks{Digital Object Identifier (DOI): see top of this page.}
}

\def\qstart{q_{start}}

\def\qgoal{q_{goal}}
\def\Qgoal{Q_{goal}}

\def\C{\mathcal{C}}
\def\CF{\mathcal{C}_{free}}
\def\R{\mathbb{R}}

\newcommand{\cspace}[0]{\mathcal{C}}
\newcommand{\sspace}[0]{\mathcal{S}}
\newcommand{\pspace}[0]{\mathcal{P}}

\newcommand{\cfree}[0]{\mathcal{C}_{free}}

\newcommand{\PREPRINTYEAR}{2025}

\newcommand{\DOI}{10.1109/LRA.2025.3540627} 
 
\usepackage[placement=top,vshift=-2,firstpage=true]{background}

\SetBgScale{1.0}

\SetBgContents{\parbox{0.95\textwidth}{\small \begin{center} The manuscript appeared in IEEE Transactions on Robotics under DOI: \DOI\end{center} \vspace{-0.2cm}  \copyright{}~\PREPRINTYEAR~IEEE.~Personal use of this material is permitted. Permission from IEEE must be obtained for all other uses, in any current or future media, including reprinting/republishing this material for advertising or promotional purposes, creating new collective works, for resale or redistribution to servers or lists, or reuse of any copyrighted component of this work in other works.} }

\SetBgColor{black}

\SetBgAngle{0}

\SetBgOpacity{1.0}

\begin{document}

\maketitle

\begin{abstract}
The asymptotically optimal version of Rapidly-exploring Random Tree (RRT*) is 
often used to find optimal paths in a high-dimensional configuration space.
The well-known issue of RRT* is its slow convergence towards the optimal solution.
A possible solution is to draw random samples only from a subset of the configuration space that
is known to contain configurations that can improve the cost of the path (omniscient set).
A fast convergence rate may be achieved by approximating the omniscient with a low-volume set.
In this paper, we propose new methods to approximate the omniscient set and methods for their effective sampling.
First, we propose to approximate the omniscient set using several (small) hyperellipsoids defined by sections of the current best solution. 
The second approach approximates the omniscient set by a convex hull computed from the current solution.
Both approaches ensure asymptotical optimality and work in a general n-dimensional configuration space.
The experiments have shown superior performance of our approaches in multiple scenarios in 3D and 6D configuration spaces. 
\\The proposed methods will be available as open-source on https://github.com/BipoaroXigen/JPL.
\end{abstract}

\begin{IEEEkeywords}
 Motion and Path Planning; Planning, Scheduling and Coordination
 \end{IEEEkeywords}

\section{Introduction}

\begin{figure}[!ht]
\centering
{\small
\setlength{\tabcolsep}{2pt}
\def\WW{0.22}
\begin{tabular}{cc}
\includegraphics[width=\WW\textwidth]{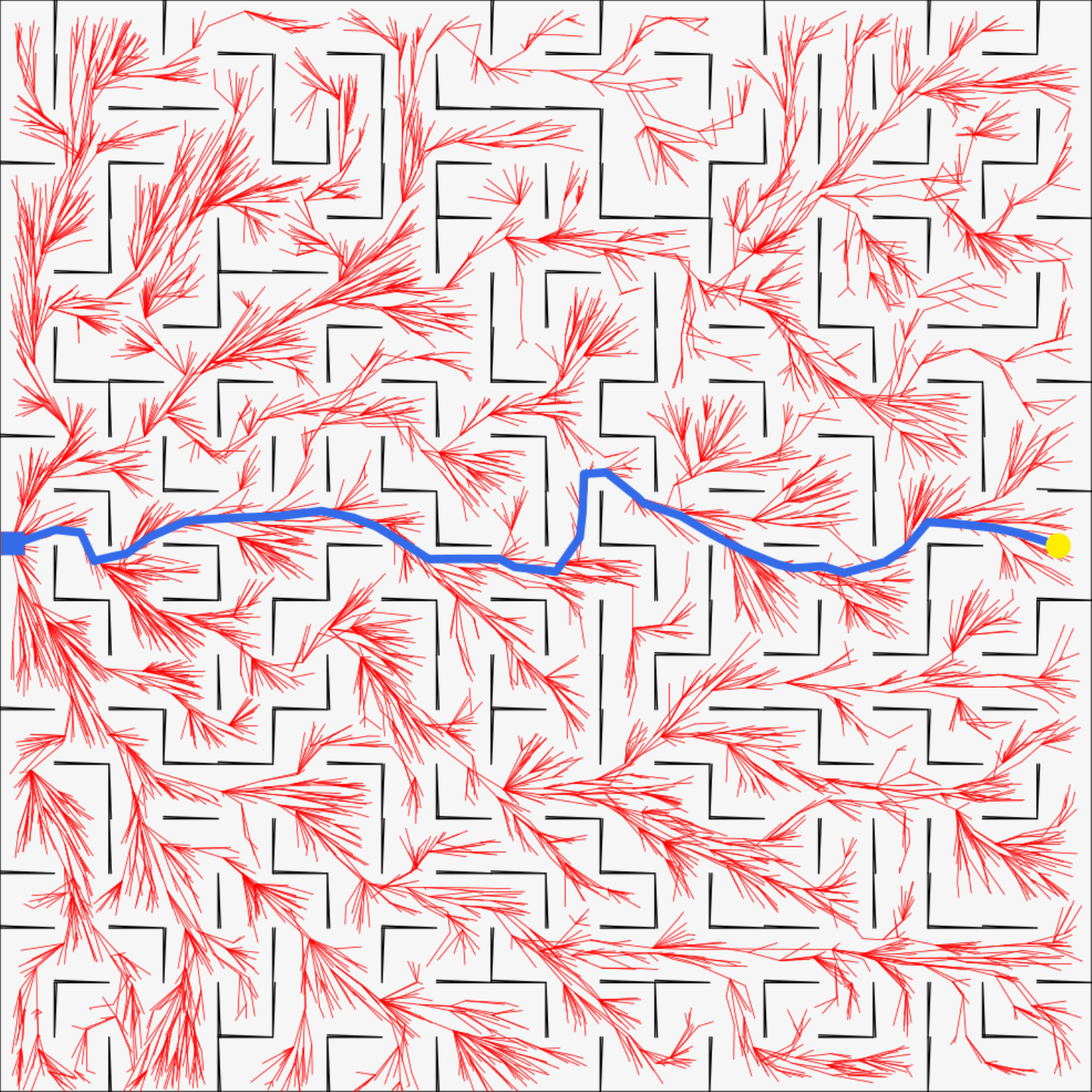} &
\includegraphics[width=\WW\textwidth]{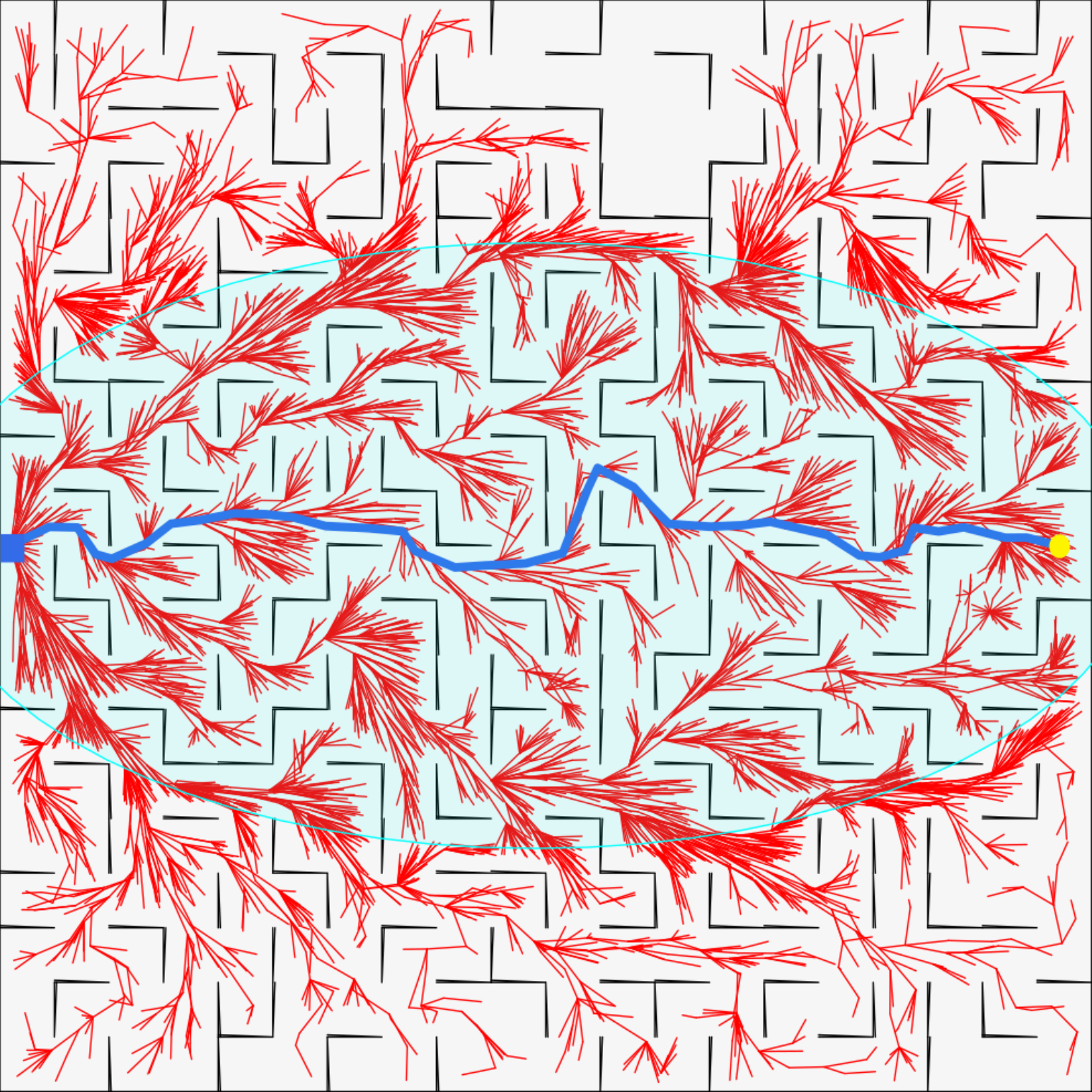} \\
RRT*, cost 563 & Informed-RRT*, cost 554 \\
\includegraphics[width=\WW\textwidth]{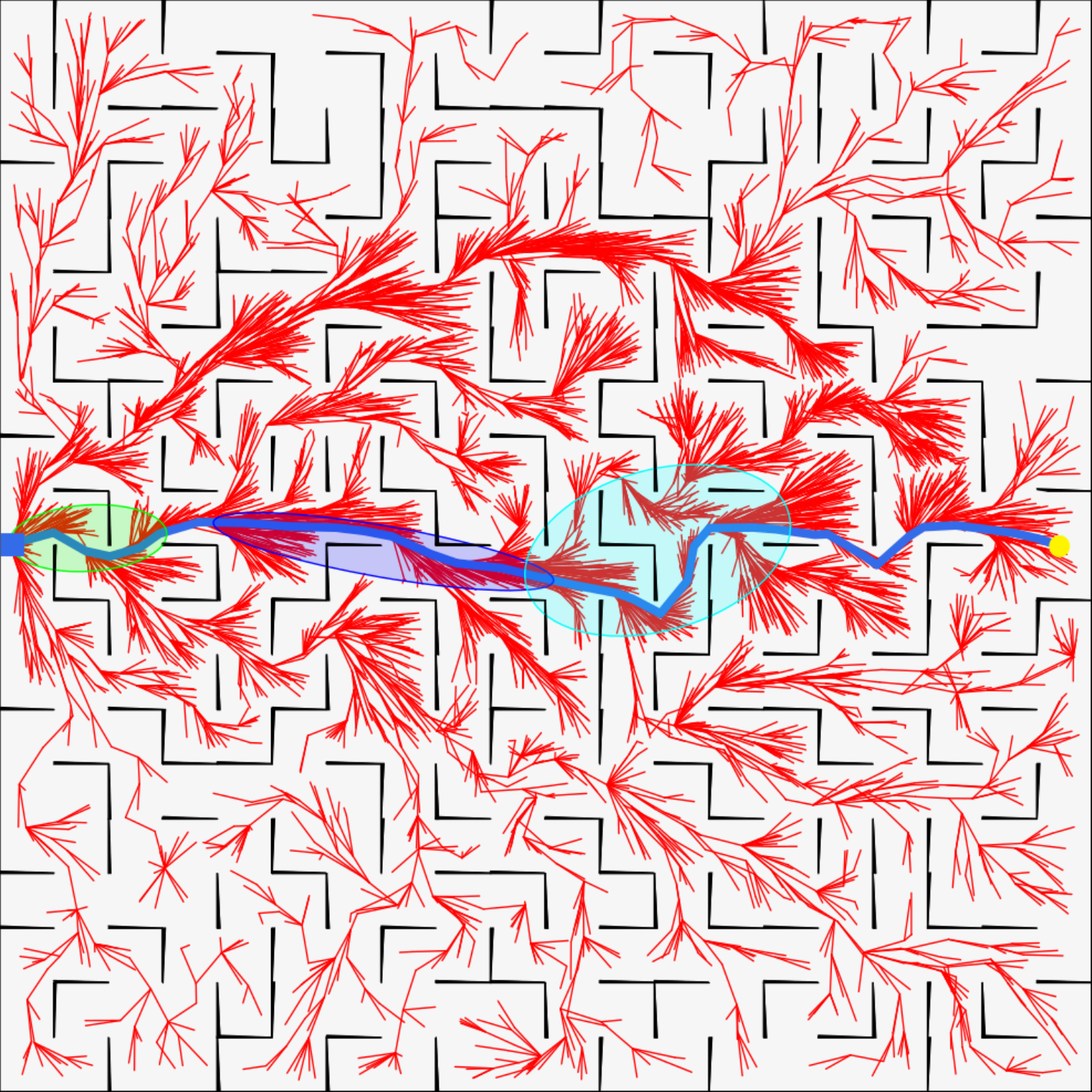} &
\includegraphics[width=\WW\textwidth]{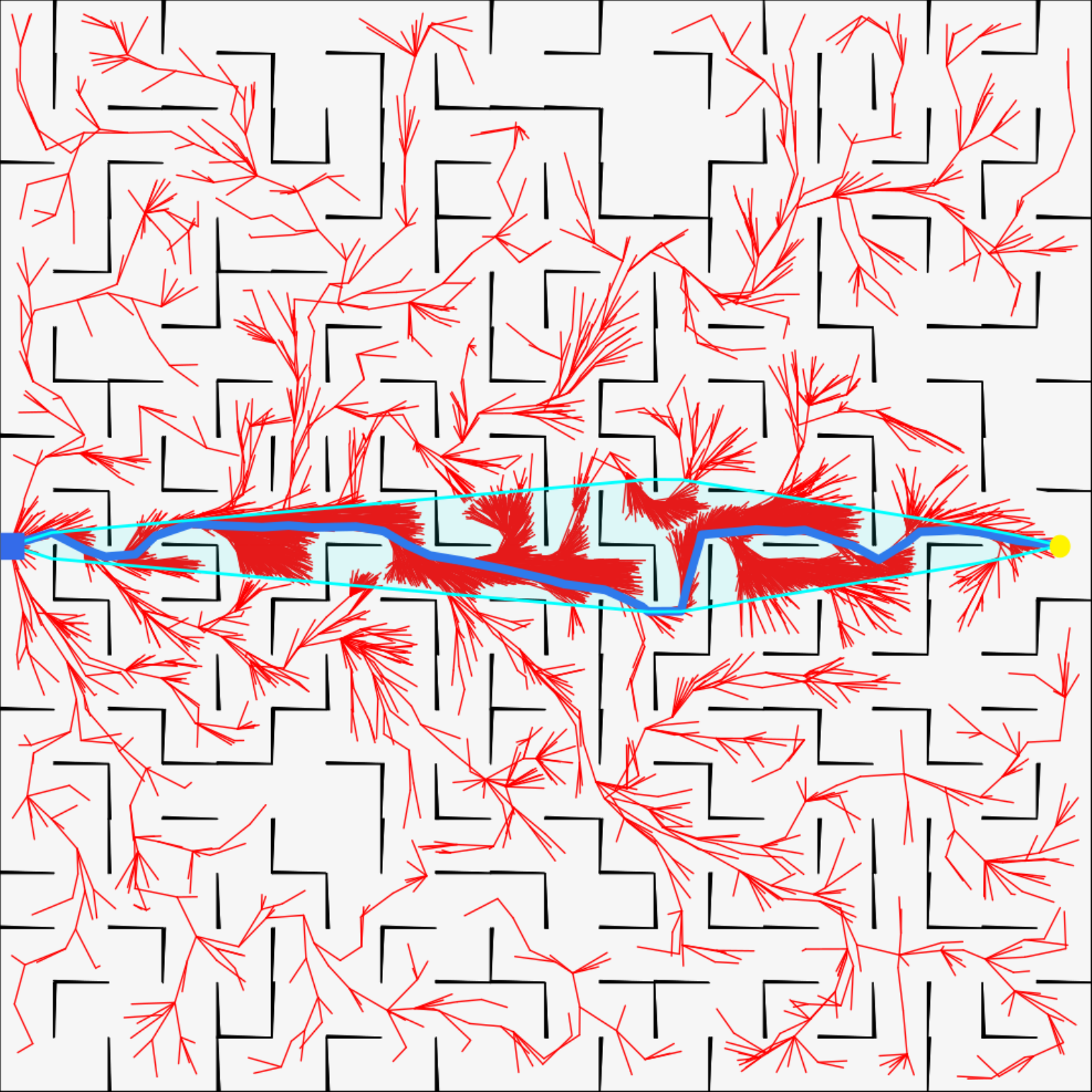} \\
PI-RRT*, cost 532 & C-RRT*, cost 531 \\
\end{tabular}
}
\captionof{figure}{
Examples of trees with $7,000$ nodes (red) with the current best solution (blue path).
RRT* samples the whole space (a), and Informed-RRT* samples from one (blue) hyperellipsoid (b).
Proposed PI-RRT* samples from multiple small hyperellipsoids (c) and
proposed C-RRT* samples from a convex set around the current best solution (d).
\label{fig:teaser}}
\vspace{-0.6cm}
\end{figure}

\IEEEPARstart{T}{he} task of optimal path planning is to find a collision-free path with the lowest cost (e.g., path length) from a start configuration to a goal configuration.
Low-dimensional configuration spaces can be discretized, and the optimal path can be searched using, e.g., A*.
Sampling-based motion planners, e.g., Rapidly-exploring Random Tree (RRT)~\cite{lavalle1998rapidly},  search the configuration space using randomized sampling, and they are
more suitable for searching high-dimensional spaces than the discretization methods.
RRT*~\cite{karaman2011sampling} is an asymptotically optimal variant of the RRT algorithm.
RRT* continues the search even after the first feasible solution is found.
Moreover, RRT* uses a rewiring technique to optimize node connection within the tree, so the costs of the nodes decrease with the increasing number of samples.
The rewiring process relies on nearest-neighbor search, and it becomes more computationally intensive with the increasing size of the tree, which leads to the well-known slow convergence of RRT*~\cite{gammel2014informed,armstrong2021rrt,gammell2015batch}.

RRT* samples the whole configuration space (Fig.~\ref{fig:teaser}a), which is not necessary, as there exist states that cannot possibly improve the existing solution.
This was first observed in~\cite{gammel2014informed} where the ``omniscient set'' is defined.
The omniscient set is a subset of the configuration space suitable for finding the optimal solution.
In obstacle-free environments, it has a form of prolate n-dimensional hyperellipsoid.
Drawing samples only from the hyperellipsoid has been shown to improve the convergence towards the optimal solution.
However, for long zig-zag paths, the volume of the hyperellipsoid may still be too large, which 
causes Informed-RRT* to perform similarly as RRT* (Fig.~\ref{fig:teaser}a,b).

In this paper, we propose two approaches to approximate the omniscient set.
The first proposed approach employs multiple small hyperellipsoids defined by subsections of the current best solution (Fig.~\ref{fig:teaser}c). 
This set ensures asymptotic optimality.
The second approach computes a convex hull of a path rotated along a line from start to goal (Fig.~\ref{fig:teaser}d).
Finally, we combine these two approaches and show how to achieve asymptotic optimality with them.
We show how to efficiently sample these sets.
Both approaches can be extended to higher dimensions.
In comparison to state-of-the-art methods, the proposed approaches converge faster towards the optimal solution.

\section{Related Work}

The family of sampling-based planners attempts to solve the path planning problem by randomized
sampling of the configuration space. 
A popular method is RRT~\cite{lavalle1998rapidly}, and its variants~\cite{verasSystematicLiteratureReview2019,elbanhawiSamplingBasedRobotMotion2014b}.
The original RRT provides a feasible solution, but it cannot guarantee finding the optimal solution.

In~\cite{karaman2011sampling}, asymptotically optimal sampling-based planners RRT* and PRM* were proposed. 
In RRT*, the connection of the tree's nodes is optimized using a rewiring procedure to achieve near-optimal solutions.
In contrast to basic RRT, where the tree growth is terminated if it reaches the goal configuration, RRT* continues to grow (and rewire) the tree even after the first feasible solution is found. 
Various improvements of RRT* were proposed:
a) methods changing the sampling distribution,
b) using multiple trees,
c) improved rewiring procedures, and d) other techniques focusing on low-level routines. The surveys~\cite{gammell2021asymptotically,verasSystematicLiteratureReview2019,orthey2023samplingbased,noreen2016optimal} cover the contributions to the field of asymptotically optimal motion planning.

{\it Adapting the sampling distribution}.
The original RRT* samples the (whole) configuration space.
As was shown in~\cite{gammell2018informed,gammel2014informed}, not all configurations can improve the quality of the existing solution.
Therefore, the sampling can be made within 
a subset of the configuration space.
In Informed-RRT*~\cite{gammel2014informed}, the authors define 
a set of configurations that can improve the path cost.
Generally, this set is an n-dimensional prolate hyperspheroid.
After the initial feasible solution is found, the random samples are generated from this subset. 
This reduces the volume of space to be sampled and improves the convergence rate.
The BIT* planner~\cite{gammell2015batch} processes the samples of the hyperspheroid in batches: random samples are generated in the given area and processed (i.e., used for tree expansion) in the order given by a chosen heuristic.
Cloud RRT*~\cite{donghyuk2014cloud} decomposes the search space by a set of spheres with
various priorities and uses the spheres to generate random samples.
Initially, the spheres are computed using a Generalized Voronoi Graph of the workspace and
updated if a new, better path is found.

The work~\cite{wang2020neural} employs a 
Convolutional Neural Network (CNN) to provide the sampling distribution. 
CNN is trained on a 2D map where optimal paths are computed using A*, but 
due to representing the map as an RGB image, the method~\cite{wang2020neural} is limited only to 2D configuration spaces.
The works~\cite{qureshiPotentialFunctionsBased2016,fan2022uav} proposed to steer the new nodes of RRT* towards the goal using the Potential Field method.
In~\cite{ichter2018learning}, the sampling distribution is learned using a variational autoencoder from a set of successful motion plans.
The learned distribution improves the convergence rate of BIT*.
The work~\cite{wang2022gmr} uses the Gaussian Mixture model to learn 
the sampling distribution from human-demonstrated trajectories.
RRT\#~\cite{arslan2013use} is a two-stage planner: exploration, which captures the topology of the space and extends the tree, and exploitation, which attempts to improve the current solution. 
Nodes having the potential to be a part of the optimal solution are estimated using available knowledge about the cost of the best path and cost-to-come values, which allows to focus the sampling to promising regions of the configuration space.

{\it RRT* with multiple trees.}
RRT*-Connect employs the bidirectional search~\cite{klemm2015rrt}: two trees are grown (from the start and goal), both in the RRT* manner, and a new node extending one tree is also tested for extending the other tree.
The work~\cite{mashayekhi2020informed} combines Informed-RRT* with the bidirectional search~\cite{mashayekhi2020informed}.
The bidirectional IB-RRT~\cite{qureshi2015intelligent} estimates the best tree (and its best node) where
to insert the random samples.
The work~\cite{strub2022adaptively} is based on BIT* and uses the bidirectional backward and forward search. 
The backward search is fast as it checks collisions approximately, and its purpose is to estimate the heuristic.
On the contrary, the forward search checks the collisions fully, grows in the BIT* manner, and uses the heuristic from the backward search.
The forward search informs the backward search about invalid (colliding) edges, so the heuristic is updated during the search.

{\it RRT* with improved rewiring}.
In RRT*, parents for the newly added node are vertices in the hypersphere around the new node.
Quick-RRT*\cite{jeong2019quick} extends this set also to the ancestors of these vertices.
F-RRT*~\cite{liao2021f} further extends this idea by creating new parent nodes instead of selecting them amongst nodes in the tree.
T-RRT*~\cite{devaurs2016optimal} extends RRT* by the transition test to prefer extension by nodes with lower costs.

{\it Improving low-level routines.}
RRT* performs a vast amount of collision checks, which can be computationally intensive.
The paper~\cite{hauser2015lazy} proposed to perform collision detection of edges only
if a new path is computed.
The work~\cite{adiyatov2017sparse} associates obstacle proximity information with each node,
and collision detection is computed only if the nearest neighbor is too far or if the node is close to obstacles.
In memory-efficient RRT*~\cite{adiyatov2013rapidly}, authors limit the maximum size of the tree, which is achieved by removing provably useless nodes every time a new node is added to the tree.
The authors of~\cite{armstrong2021rrt} extend RRT* for the purpose of online path planning.
Besides the standard Euclidean metric, the work~\cite{armstrong2021rrt} employs a second assistive metric that helps to improve the convergence rate.
An extension of the node expansion for kinodynamic systems without a BVP (Boundary Value Problem) solution was proposed in~\cite{li2016asymptotically}.
The tree is propagated using the forward system simulation. 
Moreover, the nodes are pruned based on their cost, which reduces the size of the trees and speeds up the nearest neighbor search.

In this paper, we focus on adapting the sampling space.
The most relevant work is the Informed RRT*~\cite{gammel2014informed,gammell2018informed}  
which defines the sampling space as a hyperellipsoid guaranteed to contain the optimal path.
The hyperellipsoid is parametrized by the best (shortest) path found so far.
The downside of~\cite{gammel2014informed} is that when the current shortest path is too wiggly, the hyperellipsoid may be even bigger than the whole free configuration space.
We propose several techniques to approximate the omniscient set~\cite{gammell2018informed} and to focus the sampling to more relevant regions of the configuration
space, as is demonstrated in Fig.~\ref{fig:diferences}.

\section{Problem Formulation}

Let $\C$ denote the configuration space, and let $\CF \subseteq \C$ denote the free region where the robot can move.
A path $\pspace = (p_1, p_2,\ldots,p_n),\ p_i \in \cspace$ is a sequence of configurations.
A subsection of the path is defined as a sequence $\pspace _{j,k} = (p_i),\ i=j,\dots,k,\ p_i \in \pspace$.
The length of the path $\pspace$ is denoted $len(\pspace)$, and it corresponds to the sum of distances between path elements $\sum_{i=2}^{n} ||p_{i-1}-p_{i}||$.

The task is to find the optimal (i.e., minimizing $len(\pspace)$) collision-free path $\pspace_{opt} = (p_i),\ p_i \in \CF$ from the start configuration $\qstart \in \cfree$ to the goal region $\Qgoal \subseteq \C$, i.e., 
$p_1 = \qstart$ and 
$p_n \in \Qgoal$.
In the rest of the paper, $n$ refers to the cardinality of $\pspace$.

\section{The Motivation behind Proposed Methods}

The original RRT and RRT* algorithms sample the whole configuration space $\C$ uniformly.
We define the sampling space $\sspace \subseteq \cspace$ as the region from which the random samples are drawn (i.e., in RRT and RRT*, $\sspace =\C$).
As was shown in~\cite{gammell2018informed}, only a subset of the configuration space contains samples that can possibly improve the cost of the path $\pspace$ (and it is guaranteed that samples outside this set cannot improve the cost of the path $\pspace$).
This set is called the ``omniscient\ set'' $\mathcal{O}$:
\begin{equation}
    \mathcal{O} = \{q|\ len(\pspace_{to}) + len(\pspace_{from}) \le len(\pspace);\ q \in \cspace\},
\end{equation}

\noindent
where $\pspace_{to}$ is a path from $\qstart$ to a configuration $q$ and $\pspace_{from}$ is a path from a configuration $q$ to $\qgoal$.
The lengths of the paths can be approximated with a heuristic.
The Euclidean distance heuristic would lead to the ``informed set'' $S_i$, which is a prolate hyperellipsoid
\begin{equation}
    \sspace_{i} = \{q|\ \lVert \qstart - q\rVert + \lVert q - \qgoal \rVert \le len(\pspace);\ q \in \cspace\}.
\end{equation}
\noindent

To improve the length of the path, it is sufficient to draw 
random samples only from the informed set $\sspace_i$, as no configurations outside $\sspace_i$
can improve the cost of the path.
This is the core of Informed-RRT*~\cite{gammel2014informed,gammell2018informed}
which draws random samples only from $\sspace_i$, and where the set $\sspace_i$ is 
defined using the length of the current best path.

\begin{figure}[htb]
\vspace{-0.6cm}
    \centering
    \subfloat[Sampling space $\sspace$ of Informed-RRT* for a given path.]{
	    \includegraphics[width=0.4\linewidth]{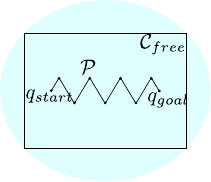}
        \label{fig:diferences_inf}
}
    \hspace{0.02\linewidth}
    \subfloat[Sampling space $\sspace$ of one of our proposed planners for the same path.]{
	    \includegraphics[width=0.4\linewidth]{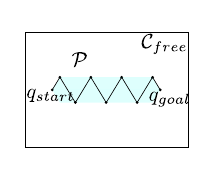}
        \label{fig:diferences_conv}
}
    \vspace{-0.15cm}
    \caption{Comparison of $\cspace_{free}$ (white) and the sampling spaces (blue).
    }
	\label{fig:diferences}
\end{figure}

However, the volume of  $\sspace_i$ can still be quite high (especially for long zig-zag paths), which is depicted in Fig.~\ref{fig:diferences}.
Moreover, it is not guaranteed that all configurations from the informed set $\sspace_i$ can improve the path.

The smaller the volume of $\sspace$, the less time will be spent sampling non-improving configurations. 
Ideally, the sampling space would be exactly the desired optimal path (i.e., $\sspace = \pspace_{opt}$). However, $\pspace_{opt}$ is not known in advance.

To reduce the number of samples that do not improve the solution, we propose several
approaches to approximate the omniscient set $\mathcal{O}$, and we propose
methods for their sampling. 

\section{Locally Informed Sampling Space}

The first proposed sampling space is a modification of $\sspace_i$ of Informed-RRT*~\cite{gammel2014informed}.
Instead of constructing a hyperellipsoid from the whole path, many smaller hyperellipsoids are constructed from various subsections of the path.
Each path subsection $\pspace _{j,k}$ has a corresponding hyperellipsoid $s_{j,k}$:
\begin{equation}
	s_{j,k} = \{q|\ (\lVert q-p_{j}\rVert + \lVert q-p_{k} \rVert) \le len(\pspace_{j,k});\ q \in \cspace;\},
\end{equation}
$s_{j,k}$ constructed in this manner is guaranteed to contain the shortest path from $p_{j}$ to $p_{k}$ as proven in the original Informed-RRT* paper~\cite{gammel2014informed}, section III.
 
The Locally Informed Sampling Space, which we will denote $\sspace_{l}$, is the union of all local hyperellipsoids $s_{j,k}$, which satisfy constraints given by the parameter $c,\ 2 \le c \le n$:
\begin{equation}\label{eq:sl_def}
	\sspace_{l} = \bigcup_{\substack{j,k \; \in \; \{1, ..., n\} \\ k > j\;,\; k - j \ge c}} s_{j,k}.
\end{equation}

\begin{figure}[h]
	\includegraphics[width=0.5\linewidth]{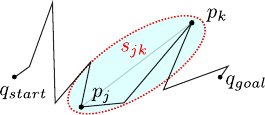}
	\centering
	\caption{Visualization of one $s_{j,k}$ in blue on the path $\pspace$.
		\label{fig:local}
		}
\end{figure}

With $\sspace _l$, the sampling approach of Informed-RRT* gets applied to parts of the current shortest found path, as can be seen in Fig~\ref{fig:local}.

The parameter $c$ controls the explore-exploit tradeoff.
With the parameter $c=n$, sampling from $\sspace_l$ is equivalent to sampling from $\sspace_i$ (i.e., in the same
manner as in the Informed-RRT* planner).
High values of $c$ lead to exploration, as the set $\sspace_l$ contains only the larger hyperellipsoids, and it 
supports the discovery of new alternative optimal solutions.
In contrast, with the low values of $c$, the set $\sspace_l$ contains more small hyperellipsoids (computed from path subsections
$\pspace_{j,k}$ of cardinality at least $c$), which leads to the exploitation of the current best solution (i.e., smoothing).
Setting the parameter $c$ high (close to the $n$ of the initially found path) can lead to a premature halt of the smoothing and the use of $\sspace_i$ on a path not yet smoothed out.

In the following subsection, we show how to efficiently sample $\sspace_l$ without explicitly constructing all the hyperellipsoids, which would be computationally very demanding.

\begin{figure*}[!ht]
\vspace{-0.4cm}
    \centering
    \subfloat[A path in three dimensions]{
    	\includegraphics[width=0.31\textwidth]{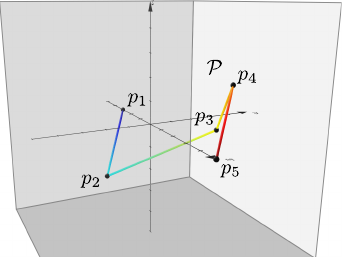}
        \label{subfig:3dpath}
}
    \subfloat[A $\sspace_c$ of path]{
        \includegraphics[width=0.31\textwidth]{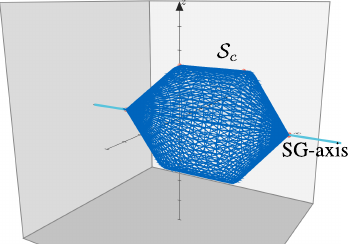}
        \label{subfig:3dsc}
}
    \subfloat[A slice of $\sspace_c$]{
        \includegraphics[width=0.31\textwidth]{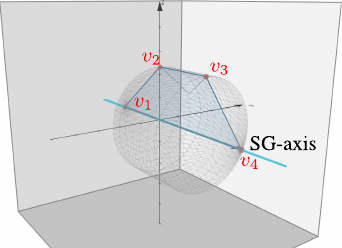}
        \label{subfig:scslice}
}
    \vspace{-0.15cm}
    \caption{\label{fig:convex3d} 
        An example of the convex sampling space $\sspace_c$ for a path 
        $P = (p_1, \ldots, p_5) = ((-3,0,0), (0,-2,-2), (2,2,0), (3,2,2), (5,0,0))$ (a).
        Sampling space $\sspace_c$ (visualized using the blue mesh) 
        is defined by $\pspace$ rotated around the SG-axis (blue) (b).
        The path can be transformed using Eq.~\ref{eq:transf} to a plane.
        The 2D convex hull of the transformed points defines the slice (blue polygon).
        The convex hull of the slice is described by $V = (v_1,\ldots, v_4)$ (c).
    }
\end{figure*}

\subsection{Drawing random samples from $\sspace_l$}
\label{sub:local_sampling}

To draw a random sample from $\sspace_l$, a random length of the subpath $\pspace_{j,k}$ is selected from range $[c,n]$, then the beginning of the path $j$ is selected randomly, the ellipsoid $s_{j,k}$ is constructed, and the random sample is generated from this ellipsoid (similarly, as Informed-RRT* does).  
The random sampling of $\sspace_l$ is summarized in Alg.~\ref{alg:loc} 
(we use symbol $U(a,b)$ for uniform sampling
in the inverval $[a,b]$).
This procedure is repeated for each new random sample.

\begin{algorithm}[!hb]
\setstretch{0.9}
\caption{Local Informed Sampling}\label{alg:loc}
\KwIn{$\pspace$: current best path; $c$: minimal length of the path segment}
\KwOut{$sample \in \sspace_l$}
\hrule
$size \gets U(c,\ |\pspace|) \in \mathbb{Z}$ \tcp*{uniform interval sample}
$j \gets U(1, |\pspace|-size) \in \mathbb{Z}$\;
$k \gets j + size$\;
$path \gets (p_j, ..., p_k)$ \tcp*{selected segment of $\pspace$}
$sample \gets informed\_sample(p_j,\ p_k,\ len(path))$\tcp*{see~\cite{gammel2014informed}, Sec. IV}
\textbf{return} $sample$\;
\end{algorithm}

\section{Convex Sampling Space\label{sec:convex}}

The second proposed sampling space is obtained
as a convex hull of revolution of $\pspace$ around the axis connecting $\qstart$ and $\qgoal$ (we refer to this axis as the SG-axis (start-goal-axis) 
in the rest of the paper) (Fig.~\ref{subfig:3dpath},~\ref{subfig:3dsc}).
We denote this sampling space as $\sspace_c$.
Computing $\sspace_c$ of the rotated path (a convex hull of an infinite set) would be complicated and unnecessary. 
Since the resulting hull is axially symmetric along the SG-axis, we can utilize that knowledge to represent $\sspace_c$ by a two-dimensional slice.

We can define a ``slice'' of the convex hull, which is an intersection of $\sspace_c $ and a plane going through the SG-axis (Fig.~\ref{subfig:scslice}).
Such a slice is two-dimensional, allowing us to represent $\sspace_c$ in 2D.
A point inside the slice can be represented by the distance along the SG-axis and the distance from the axis to the point.
Defining the slice and its coordinate system enables us to generate
random points inside $\sspace_c$ by first drawing a random sample inside the slice and then distributing the sample into the volume of $\sspace_c$.

Let $A$ be the orthogonal projection matrix onto the SG-axis.
For a configuration $q \in \C$, we define the distance along the SG-axis $a(q)$ and the distance from the axis $f(q)$ as
\begin{equation}\label{eq:dists}
    a(q) = \lVert o - Aq \rVert, \quad f(q) = \lVert q - Aq \rVert,
\end{equation}
introducing a transformation function $transf : \cspace \to \R^2$
\begin{equation}\label{eq:transf}
    transf(q) = (a(q),\ f(q)).
\end{equation}

\begin{figure}
\includegraphics[width=0.8\linewidth]{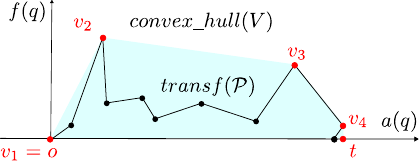}
	\centering
	\caption{The $\sspace_c$ transformed with the function $transf$ (Eq.~\ref{eq:transf}) is equal to the convex hull of the set $V$, which is easier to compute. The $transf(\pspace)$ is a transformation of the path from the Fig.~\ref{fig:local}
		\label{fig:repres}
		}
  \vspace{-0.4cm}
\end{figure}

\subsection{Slice computation \label{sec:slicecomputation}}

We project each configuration $p_i$ of the path $\pspace$ onto the SG-axis using $A$ and denote the first and the last projected configurations $o$ and $t$, respectively (the projections are ordered by their scalar projection on the SG-axis, i.e., according to their distance $a(\cdot)$ along the axis).
Note that $transf(o) = (0,0)$. 
The coordinate system of the slice is depicted in Fig.~\ref{fig:repres}.

{\bf Computation of the slice}.
Let $transf(\pspace \cup \{o,t\})$ denote a set of 2D points obtained by applying the transformation to each point of the path $\pspace$ and to the points $o$ and $t$.
As all these points now lie on a 2D plane, we can compute their 2D convex hull and obtain the set of extremal points of the hull that we denote $V$. 
The points $v_i \in V$ then define the shape (polygon) of the slice, and the whole $\sspace_c$ would be achieved by rotating this polygon around the SG-axis.
Computing the 2D convex hull of $m$ points (here, $m= n + 2$, i.e., 
number of waypoints plus two points $o$ and $t$) has
time complexity $\mathcal{O}(m \log m)$.
Practically, the method can be slightly sped up.

{\bf Efficient computation of the slice.}
The efficient computation of the convex hull of the slice points is based on a modified Graham scan~\cite{GRAHAM1972132}.
Graham scan decides whether the point lies within the hull (and therefore can not be an extremal point) by checking whether three consecutive points form a right or a left turn.

We modify the Graham scan using prior knowledge about the resulting hull
of the set of points $V=transf(\pspace \cup \{o,t\})$.
First, one edge of the convex hull is already known (it is the line segment $\overline{o,t}$).
Second, all points are located only in one direction from this edge (all points of the slice have a positive $f(\cdot)$ value).

With the mentioned constraints, we can simplify the Graham scan as follows.
Let $V=transf(\pspace \cup \{o,t\})$ and we sort points in $V$ by their $a(\cdot)$ values.
As the set $V$ is sorted, we can define the previous point $v_p \in V$ and
the following point $v_f \in V$ for a given point $v \in V$.

We can use the already known distance from the known edge (the $f(\cdot)$ value) and omit from $V$ all such points $v$ that are located between the known edge (SG-axis) 
and the line segment $\overline{v_p,v_f}$ since they lie inside the convex hull and can not be extremal.
The algorithm for omitting non-extremal points of the two-dimensional convex hull is listed in Alg.~\ref{alg:scan} 
and the process of deciding if a single point can be extremal is in Alg.~\ref{alg:in}.

\begin{algorithm}
\setstretch{0.9}
\caption{Graham scan\label{alg:scan}}
\KwIn{$V=transf(\pspace \cup \{o,t\})$}
\KwData{SG-axis}
\KwOut{$V$}
\hrule
    $V \gets \mbox{sort}\ V \mbox{ by scalar projection on SG-axis}$\;
    $i \gets 2$ \tcp*{start processing the first triplet}
	\While{True}{
		\If{inside$\_$hull$(v_{i-1},\ v_i,\ v_{i+1})$}{
                $V \gets V \backslash \{v_i\}$ \tcp*{Not extremal}
			$i \gets 2$ \tcp*{go back to the start}
			\textbf{continue}\;
		}	
		\If{$i=|V|-1$}{
			\textbf{return} $V$\;
		}
		$i \gets i + 1$\;
	}
\end{algorithm}
\vspace{-1em}

\vspace{-1em}
\begin{algorithm}
\setstretch{0.9}
\caption{Inside hull\label{alg:in}}
\KwIn{$v_p,\ v_{query},\ v_f \in \R^2$}
\KwData{$\qstart$ start configuration; $\qgoal$ goal configuration; SG-axis}
\KwOut{$Boolean$}
\hrule
	$l \gets line\ segment\ from\ v_p\ to\ v_f$\;
	$SG \gets line\ segment\ from\ \qstart\ to\ \qgoal$\;
    \If{$v_{query}$ is between $SG$ and $l$}{
		\textbf{return} $True$\tcp*{Inside}
 }
 \Else{
		\textbf{return} $False$\tcp*{Not inside}
 }
\end{algorithm}
\vspace{-2em}

\subsection{Inlier query}\label{sub:inlier}

To check if a configuration $q \in \C$ lies inside $\sspace_c$,
we construct set $R = \{transf(q)\} \cup V$ and order the elements of $R$ by their scalar projection into the SG-axis (i.e., 
according to their distance $a(\cdot)$ along the axis).
Then we proceed to find $v_p~\in~R$ and $v_f~\in~R$ for the query configuration $q$.
With these points, we can use Alg.~\ref{alg:in} to decide whether $transf(q)$ lies inside the slice of $\sspace_c$.
If $transf(q)$ lies inside the slice, then $q \in \sspace_c$.

\subsection{Drawing random samples from $\sspace_c$\label{sec:sampl}}

We can sample the set $\sspace _c$ either directly or with rejection sampling.
The rejection sampling approach is simple to implement but less efficient for high dimensional $\cspace$ or if the volume of $\sspace_c$ is low (in comparison
to the volume of the whole $\C$).
Sampling $\sspace_c$ in higher dimensions (or when the volume
of $\sspace_c$ is low) can be efficiently achieved using the direct sampling.

{\it Rejection Sampling of $\sspace_c$}.
Let $r \in \C$ be a random sample from $\C$, and $v' = transf(r)$.
We accept the sample $r$ as being in $\sspace_c$ if the point $v'$ is located
inside the 2D convex hull of the slice.

\begin{figure}[h]
\vspace{-1em}
    \includegraphics[width=0.8\linewidth]{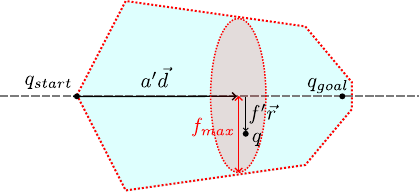}
    \centering
    \caption{
    Direct sampling of the set $\sspace_c$.
    \label{fig:sampl}
    }
\vspace{-0.4cm}
\end{figure}

{\it Direct sampling of $\sspace_c$}.
First sample $a'$ from interval $[0,\ \lVert o - t \rVert]$ 
(see definitions in subsection~\ref{sec:slicecomputation}), 
the sampling should be weighted by $f_{max}$ ($2f_{max}$ is the width of the slice of $\sspace_c$ for a given value of $a(\cdot)$) at each $a'$.
The sampled value of $a'$ determines the maximal value $f_{max}$ that $f'$ can obtain.
Then, sample $f'$ uniformly from the interval $[0,\ f_{max}]$.
This forms a random sample 
$g = (a',\ f')$.
The random configuration $q \in \cspace$ is computed as:
$q = a' \vec{d} + f' \vec{r}$, where 
$\vec{d}$ is a unit vector in direction from $\qstart$ to $\qgoal$, 
and 
$\vec{r}$ is a random unit vector perpendicular to $\vec{d}$.
The reconstruction process is illustrated in Fig.~\ref{fig:sampl}.

\section{Locally Informed Convex Sampling Space\label{sec:combination}}

\begin{figure}[h]
	\includegraphics[width=0.8\linewidth]{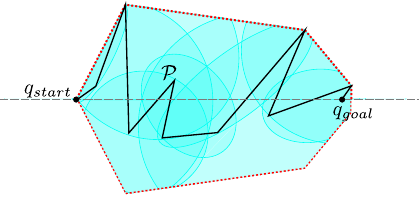}
	\centering
	\caption{An intersection of $\sspace _{c}$ and $\sspace _l$, denoted as $\sspace_{cl}$. 
		\label{fig:combination}
		}
\vspace{-0.4cm}
\end{figure}

The previously defined sampling spaces $\sspace_l$ and $\sspace_c$ can be combined; this
leads to another sampling space $\sspace_{cl} = \sspace_l \cap \sspace_c$.
While $\sspace_{c}$ will bring down the volume of the sampling space, $\sspace_{l}$ will put more weight on sampling in the proximity of the found path, locally smoothing out the fast converging solution.
Example of the space $\sspace_{cl}$ is depicted in Fig.~\ref{fig:combination}.

\subsection{Drawing random samples from $\sspace_{cl}$}
\label{sec:space_scl}
To draw a random sample from $\sspace_{cl}$, first a random
sample is generated in $\sspace_l$ (which is described in section \ref{sub:local_sampling}),
and the sample is accepted only if it is also located in $\sspace_c$ (which is realized
using the Alg.~\ref{alg:in}).

\section{Discussion}

The proposed approximations of the omniscient set and methods for their sampling can be 
integrated into any RRT*-based planner (instead of drawing random
samples from the whole $\cspace$, the planner draws random samples from $\sspace_l$, $\sspace_c$ or $\sspace_{cl}$).
Similarly to Informed-RRT*, where the set $\sspace_i$ is updated when
a new (shorter) path is found, the proposed sets 
$\sspace_l$, $\sspace_c$, and $\sspace_{cl}$ can be updated every time a new path is found.

Sampling from $\sspace_l$ (Section~\ref{sub:local_sampling})
is computationally not demanding. 
It only requires selecting a subsection of the current best solution and defining the hyperellipsoid using its first and last configurations.
Therefore, sampling from $\sspace_l$ has the same complexity  $\mathcal{O}(1)$ as sampling in Informed-RRT*.

Drawing random samples from $\sspace_c$ (and also from $\sspace_{cl}$) is computationally more demanding.
Every time a new solution is found, it is required to apply transformation in Eq.~\ref{eq:transf} to all points of the current best solution $\pspace$ and to compute the convex hull as described in section~\ref{sec:sampl}.
The time complexity of the convex hull computation is  $\mathcal{O}(n \log n)$ for a path $\pspace$ with $n$ waypoints.
More frequent hull reconstruction will result in a faster decrease of $\sspace_c$ volume. 
However, reconstructing the hull too often can slow down the planning, without much improving the convergence.
We observed that it is not necessary to construct $\sspace_c$ every time the current best solution is improved, but it is satisfactory to update it in every $m$ iterations
(in our experiments, we used $m=1,000$).

The locally informed sampling space $\sspace_l$ enables more frequent sampling close to the current best-known path.
The planners using $\sspace_l$ tend to have a faster convergence rate than ones using $\sspace_c$ in environments where less topologically distinct paths are present (for example, the Hard, Comb, and 3Dcomb environments) and slower in the opposite case.
The $\sspace_{cl}$ can be used as a compromise when there is not enough information about the environment.
Sampling from $\sspace_l$ preserves asymptotic optimality.

\textit{Proof:} 
The equation~(\ref{eq:sl_def}) is satisfied by all subpaths of cardinality from the interval $[c, n]$.
Therefore, there are $n-c+1$ possible cardinalities of subpaths, including
the whole path with the cardinality $n$.
When generating a random sample from $\sspace_l$, we first randomly (uniformly) select
a cardinality from the interval $[c,n]$ (Alg.~\ref{alg:loc}).
Therefore, with the probability $\frac{1}{n-c+1}$, we select the subpath of the cardinality $n$.
The hyperellipsoid defined by $\pspace$ is $\sspace_i$.
Therefore, when sampling from $\sspace_l$, we sample from $\sspace_i$ with probability $\frac{1}{n-c+1}$.
In~\cite{gammel2014informed} section III, it was proven that sampling of $\sspace_i$ leads to asymptotically optimal planning.
Since sampling of $\sspace_i$ guarantees asymptotic optimality and we perform it with nonzero probability, sampling of $\sspace_l$ also guarantees asymptotic optimality.
$\hfill \blacksquare$

On the contrary, the spaces $\sspace_c$ and $\sspace_{cl}$ do not guarantee to fully cover the omniscient set as $\sspace_i$ and $\sspace_l$ do, and drawing random samples only from these would result in a planner that does not ensure asymptotic optimality.
Therefore, to ensure asymptotic optimality when using $\sspace_c$ or $\sspace_{cl}$, random samples should also be generated from $\sspace_i$ with a non-zero probability.
This is an often adopted trick that combines Informed-RRT* (which samples from $\sspace_i$) with other methods because the combination preserves asymptotic optimality.

\begin{figure}[htb]
\vspace{-0.4cm}
    \centering
    \subfloat[$7,000$ iterations]{
    	\includegraphics[width=0.45\linewidth]{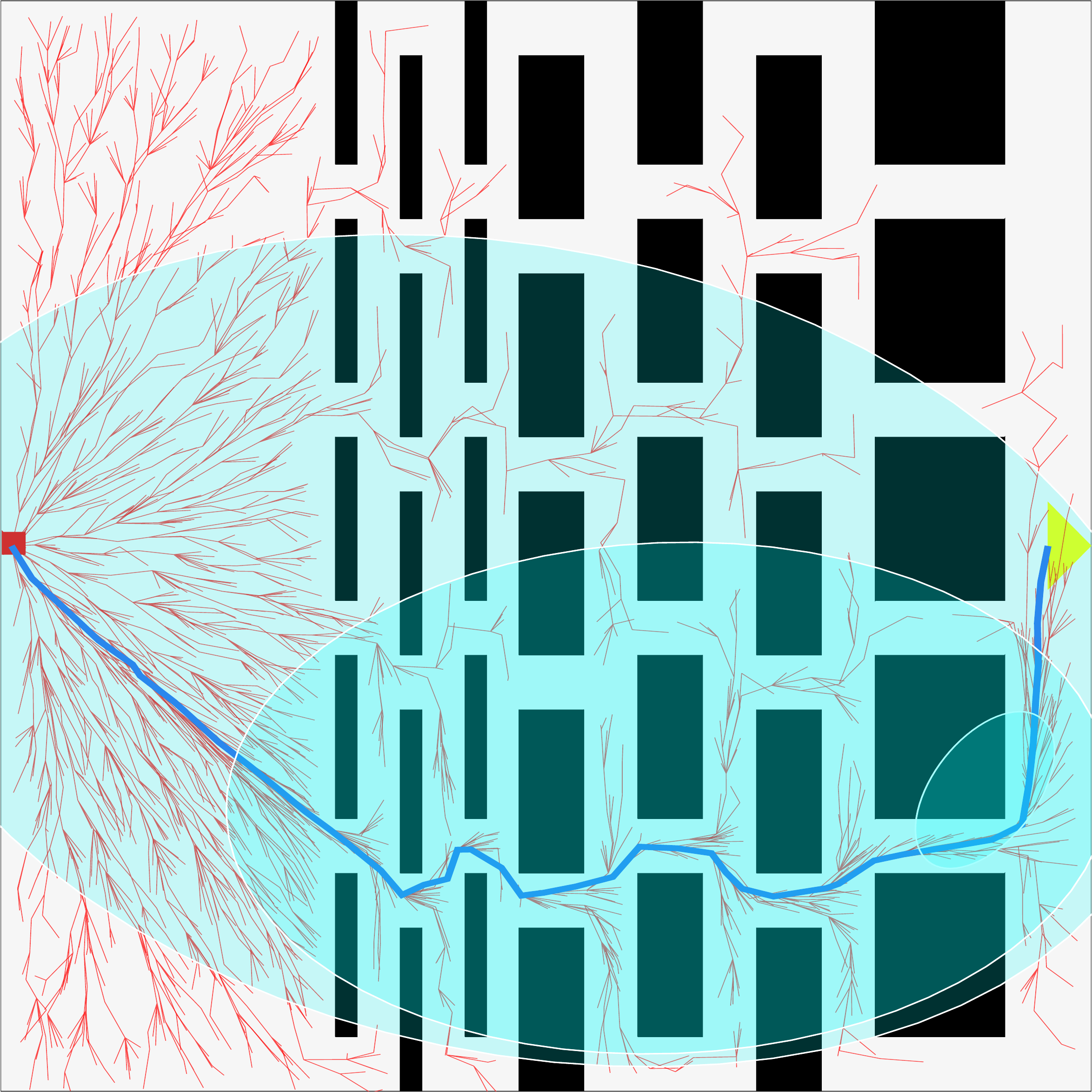}
}
    \subfloat[$14,000$ iterations]{
    	\includegraphics[width=0.45\linewidth]{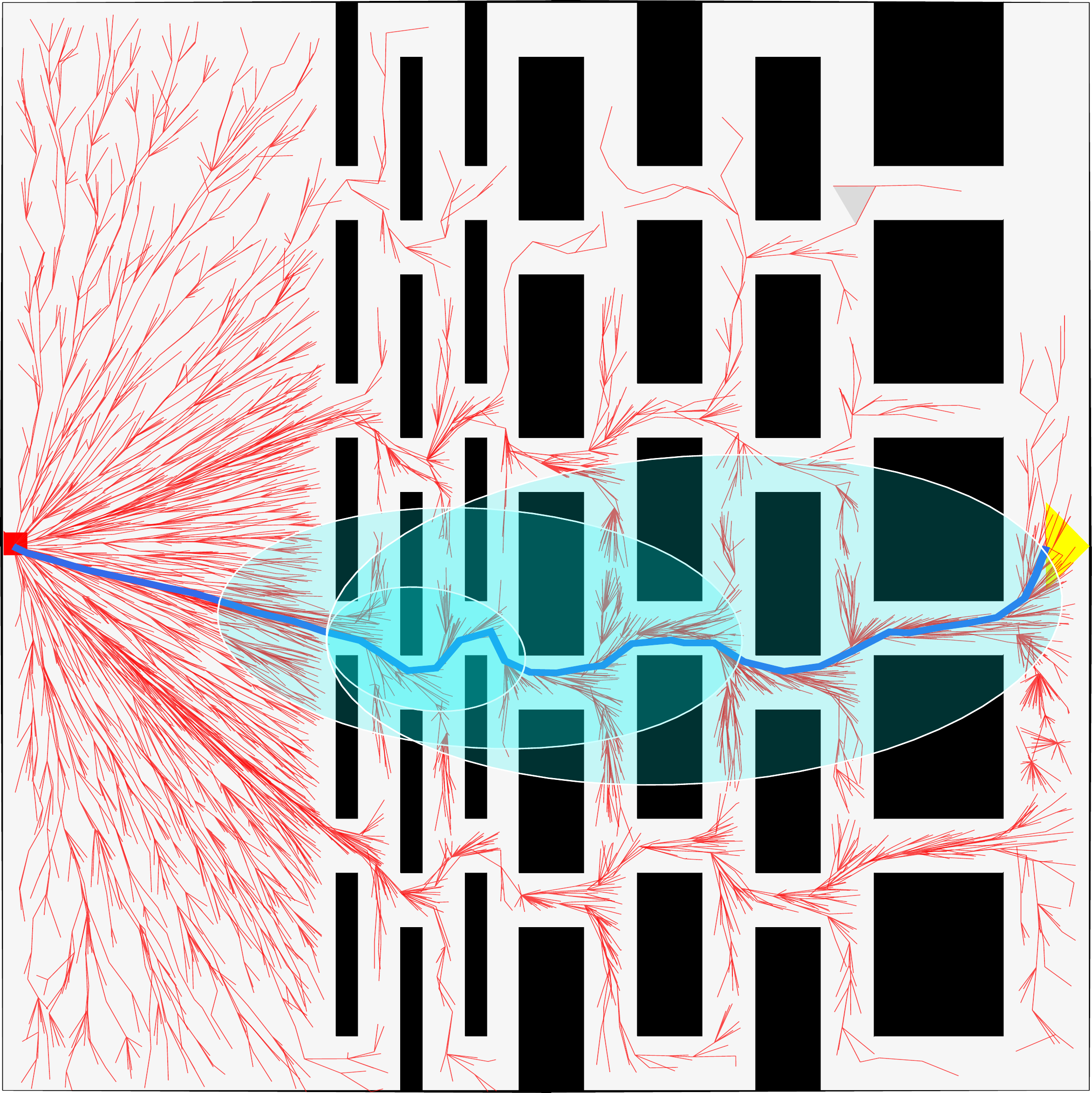}
}
\vskip 0.1pt
    \subfloat[$7,000$ iterations]{
    	\includegraphics[width=0.45\linewidth]{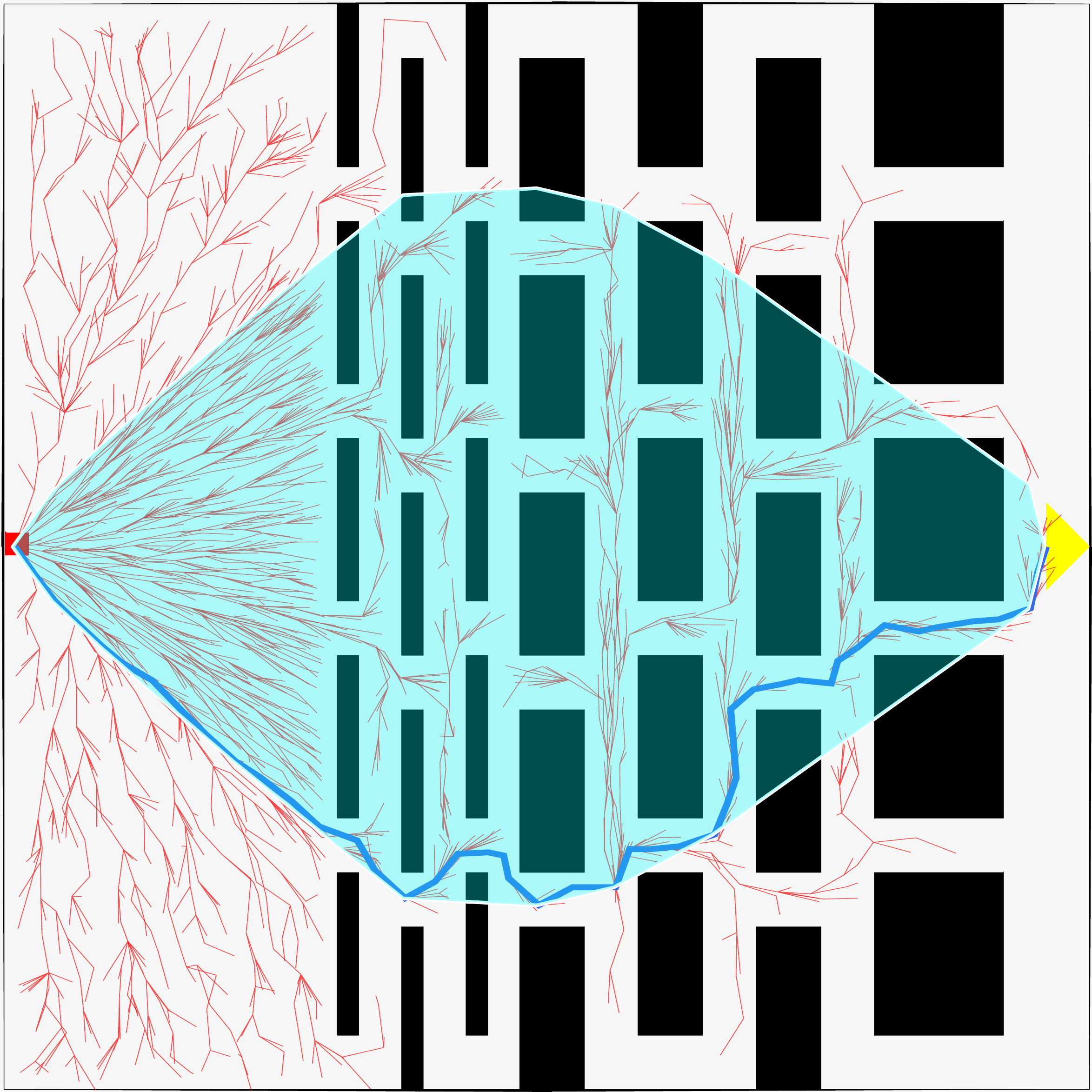}
}
    \subfloat[$14,000$ iterations]{
    	\includegraphics[width=0.45\linewidth]{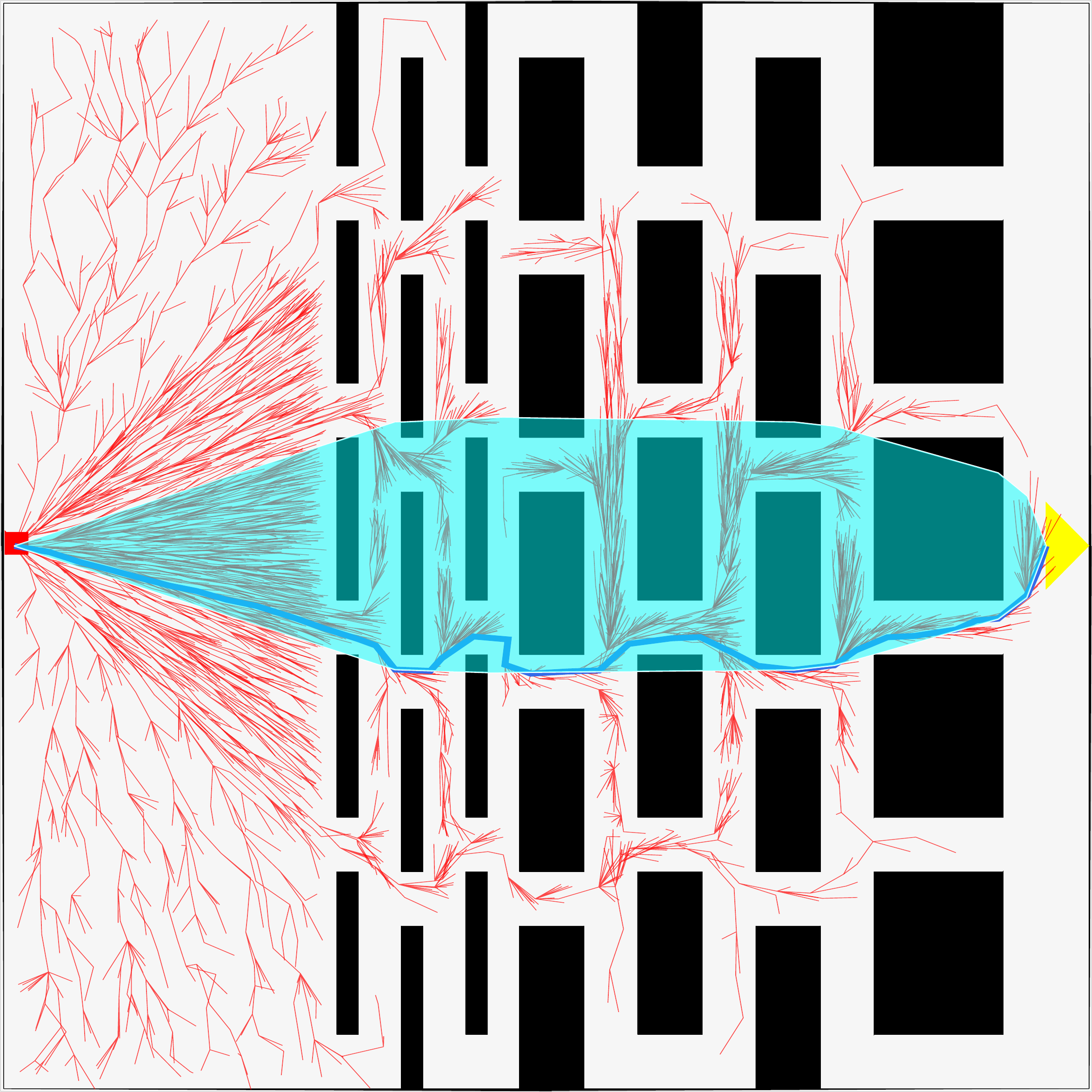}
}
    \vspace{-0.15cm}
	\caption{The tree built using PI-RRT* and C-RRT* (sampling from $\sspace_c$), the set $\sspace_c$ is in blue.
		\label{fig:search}
		}
\end{figure}

\newlength{\textfloatsepsave} 
\setlength{\textfloatsepsave}{\textfloatsep} 
\setlength{\textfloatsep}{0pt}

\section{Experiments and Results}

The proposed approximations of the omniscient set (and their sampling) were implemented and integrated
in state-of-the-art planners.
A method with the prefix `PI-' (partially informed) generates the random samples from $\sspace_l$ 
(Section~\ref{sub:local_sampling}),
the prefix `C-' (convex) denotes sampling from the convex hull $\sspace_c$ (Section~\ref{sec:sampl}), 
and finally, the prefix `PIC-' (partially informed convex) denotes the combination of `PI-' and `C-' methods $\sspace_{cl}$
(the procedure of generating the samples for `PIC-' planners is described in~\ref{sec:space_scl}).
We integrated our approaches into RRT* and BIT* methods (e.g., PI-RRT* is the RRT* planner that generates the samples from $\sspace_l$).

\begin{figure}[htb]
\vspace{-0.4cm}
    \centering
    \subfloat[Planner stuck in the upper homotopy class]{
    	\includegraphics[width=0.3\linewidth]{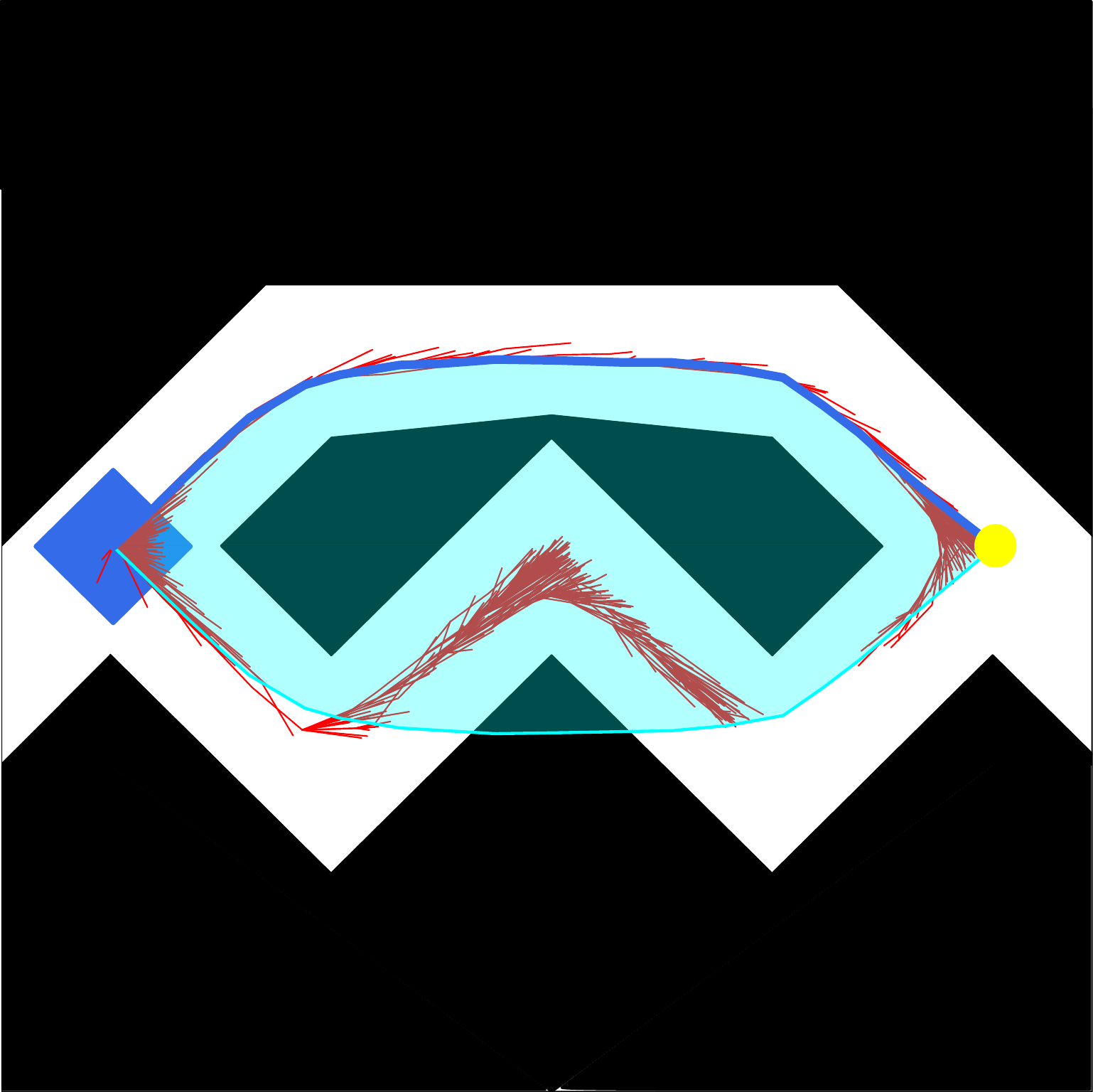}
        \centering
        \label{fig:hard_up}
}\hspace{20pt}
    \subfloat[Planner stuck in the lower homotopy class]{
    	\includegraphics[width=0.3\linewidth]{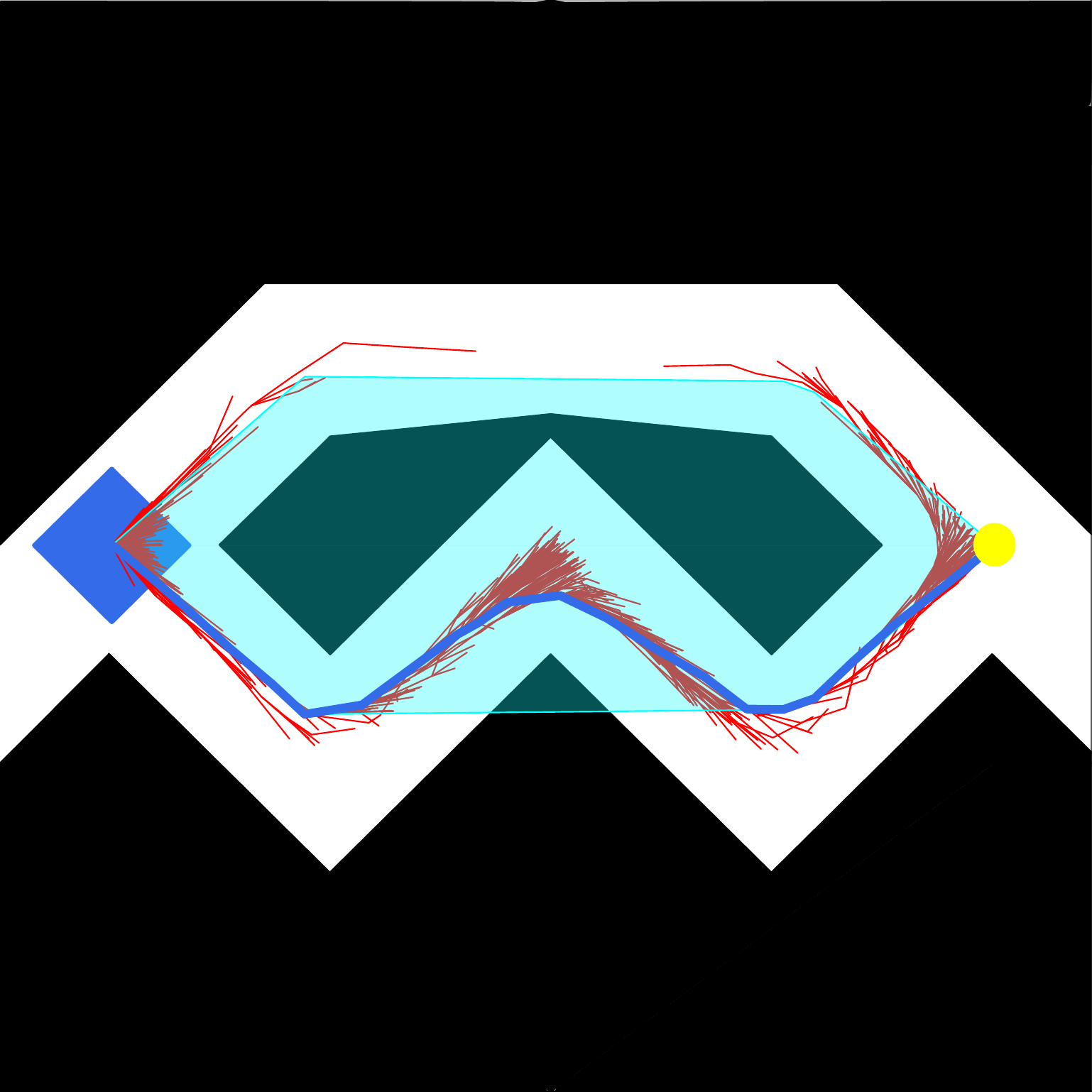}
        \centering
        \label{fig:hard_down}
}
    \vspace{-0.15cm}
    \caption{The design of the Hard environment.}
	\label{fig:2dhard}
\vspace{-0.4cm}
\end{figure}

\begin{figure}[htb]
    \centering
    \subfloat[Comb]{
    	\includegraphics[width=0.2\linewidth]{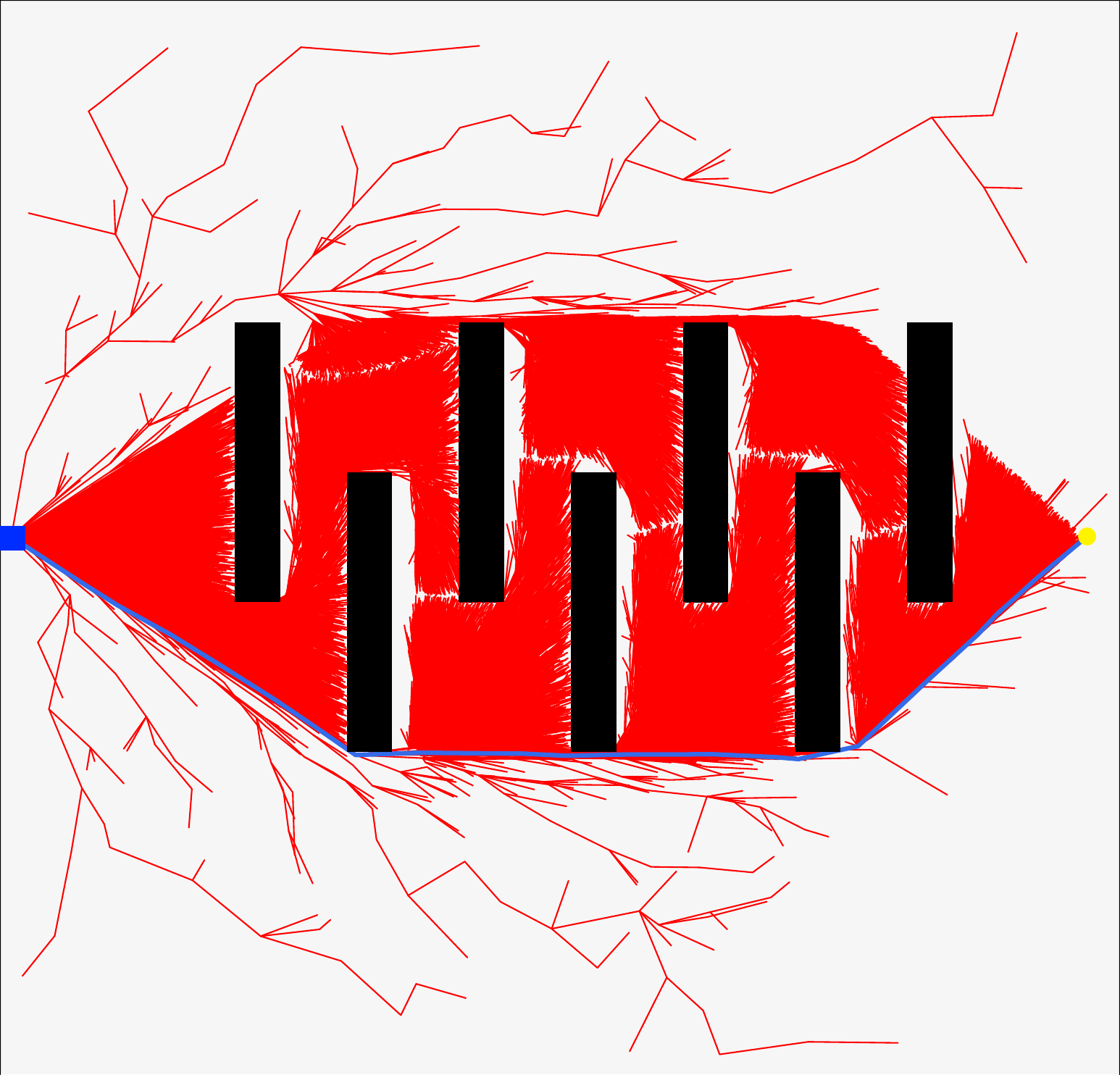}
        \label{fig:comb_tree}
}
    \subfloat[Hard]{
    	\includegraphics[width=0.2\linewidth]{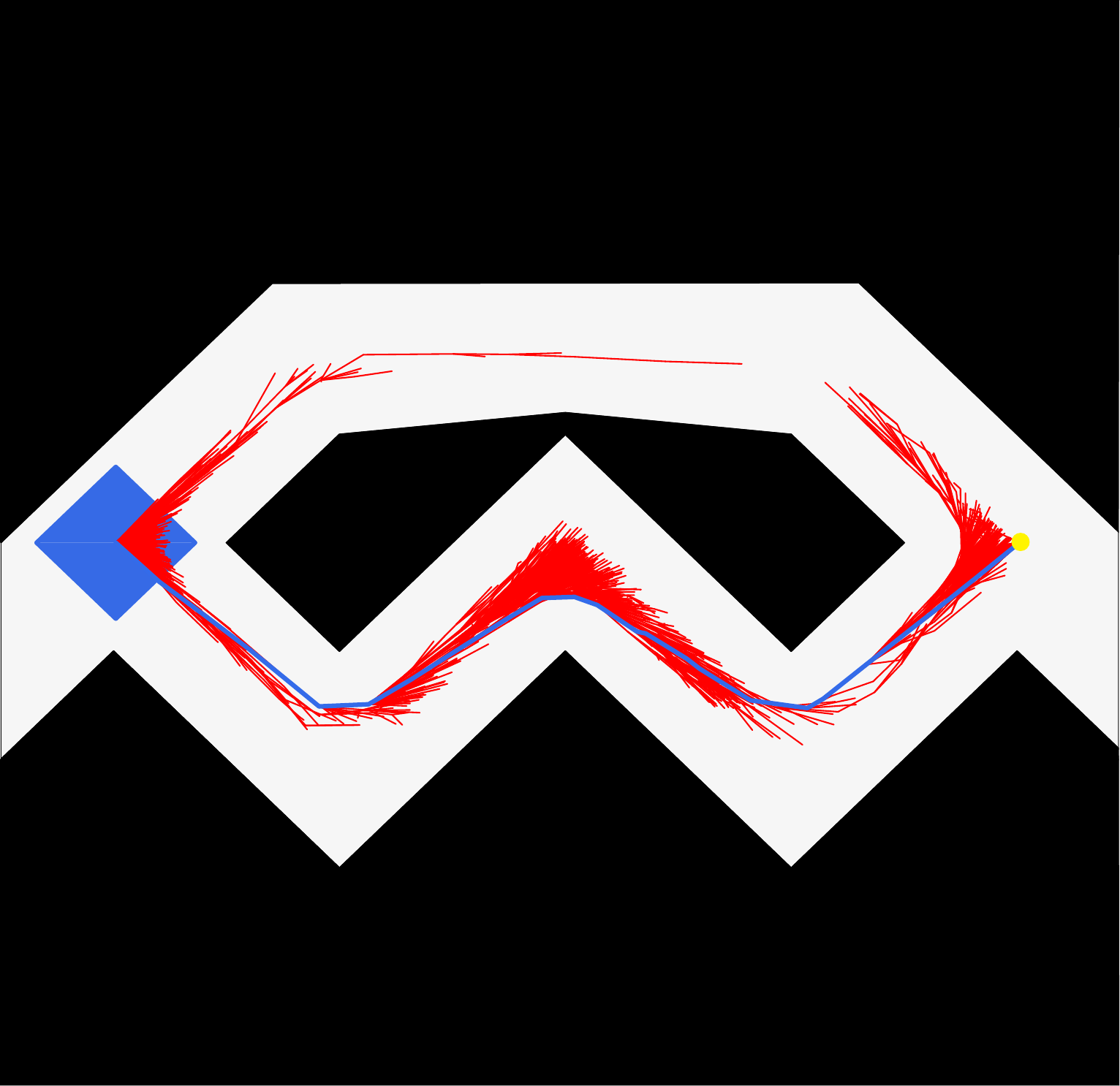}
        \label{fig:hard_tree}
}
    \subfloat[Wall]{
    	\includegraphics[width=0.2\linewidth]{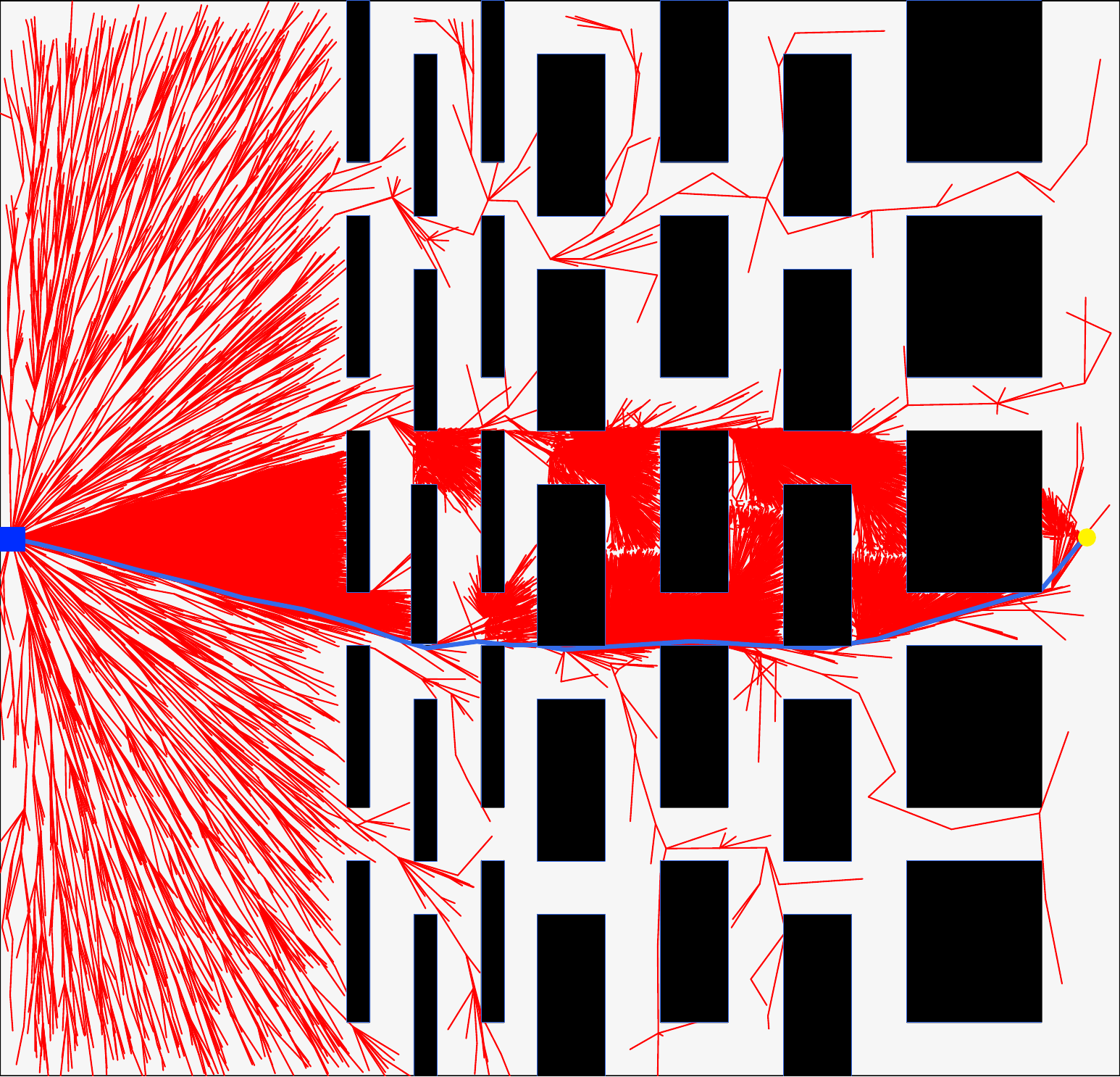}
        \label{fig:wall_tree}
}
    \subfloat[Maze]{
    	\includegraphics[width=0.2\linewidth]{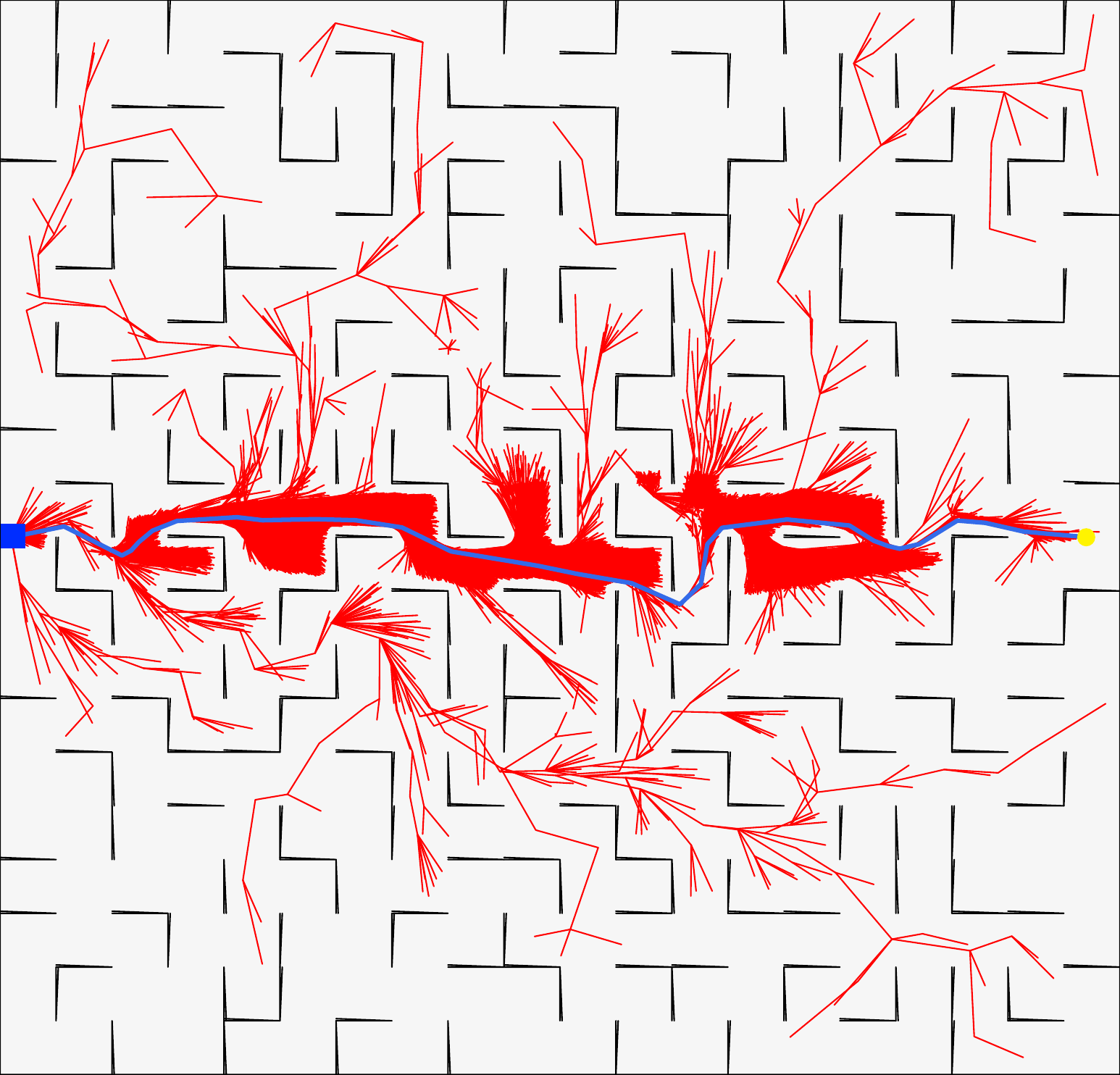}
        \label{fig:maze_tree}
}
    \vspace{-0.15cm}
    \caption{The two-dimensional environments used, with a search tree and path generated using $\sspace_{c}$.}
	\label{fig:2denvs}
 \vspace{-0.5cm}
\end{figure}

The methods were tested on path planning for a rectangle object in four 2D environments  (Comb, Hard, Wall, Maze) of size $500\times500$ (Fig.~\ref{fig:2denvs}), i.e., in 3D configuration space as the robot can translate and rotate.
The size of the object is $10\times10$ units in Comb, Wall, and Maze environments.
The Hard environment (Fig.~\ref{fig:hard_tree}) was designed specifically to pose a challenge to methods utilizing $\sspace_c$.
In this case, the robot size is $50\times50$, and the environment
contains two distinct homotopy classes (Fig.~\ref{fig:2dhard}): one
is the bottom `zig-zag' path, and the other one is the top path.
When the convex set $\sspace_c$ is computed from either path, it will not fully cover the other path.
Therefore, the `C-' planners should have a worse average performance in this case.

\begin{figure}[t]
    \centering
    \subfloat[Random]{
	    \includegraphics[width=0.45\linewidth]{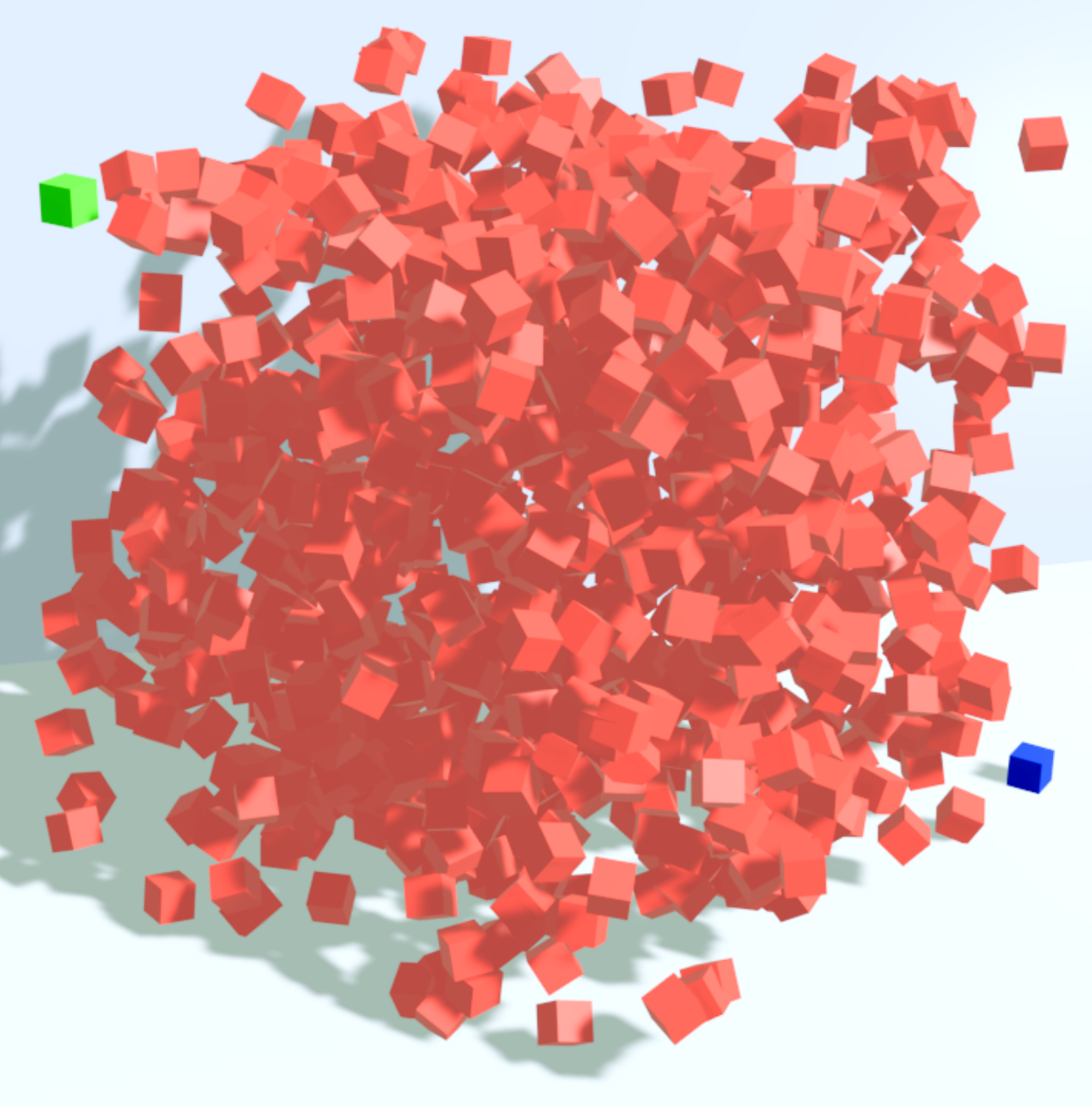}
        \label{fig:random_env}
}
    \subfloat[3D Comb]{
	    \includegraphics[width=0.45\linewidth]{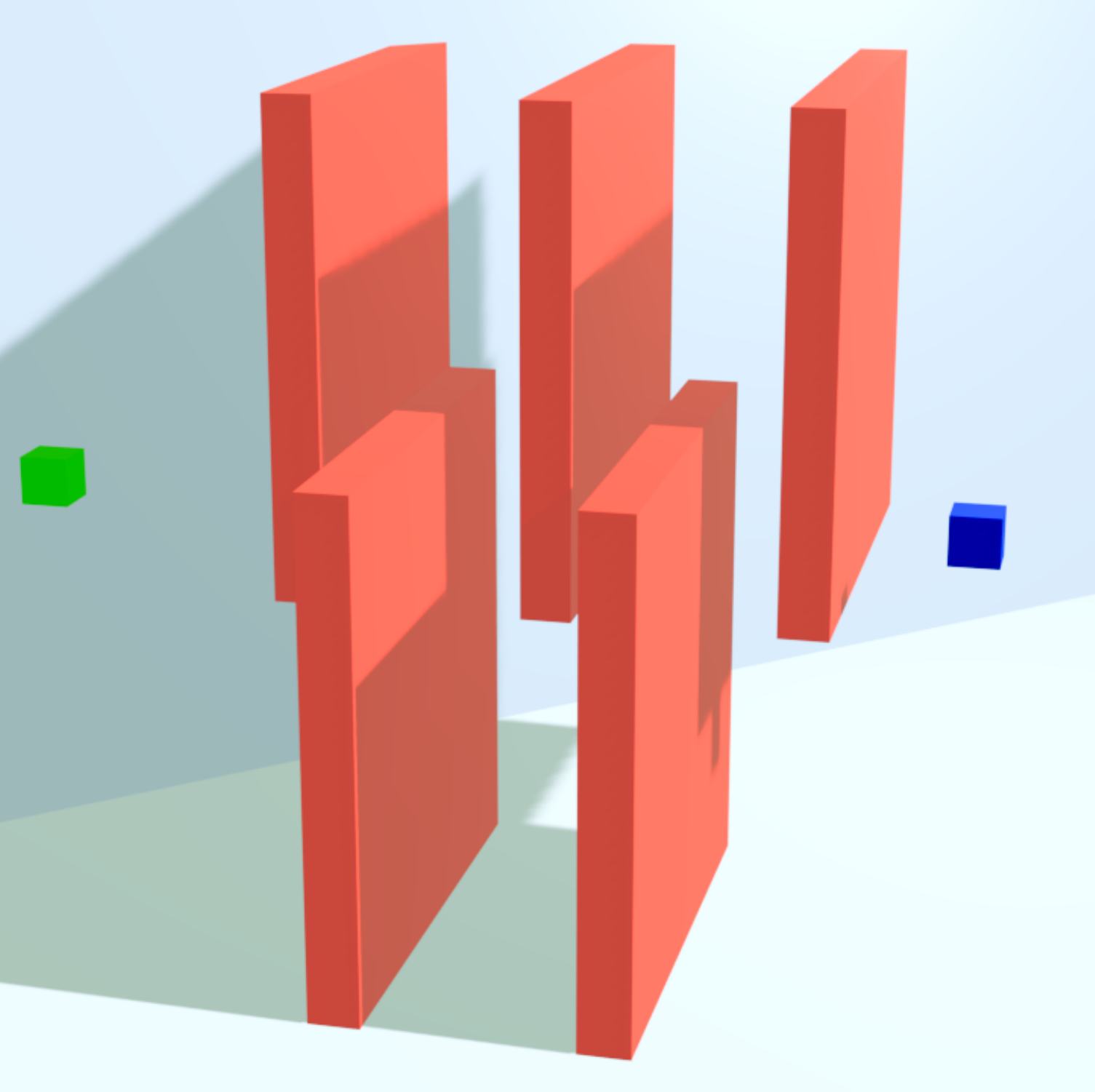}
        \label{fig:3Dcomb_env}
}
    \vspace{-0.15cm}
    \caption{3D environments, obstacles are in red, the goal region in green, and the robot (in $\qstart$) in blue.
\label{fig:3denvs}
}
\end{figure}

We also tested the performance in two 3D environments: 
Random (size $215\times215\times215)$, which is cluttered with
many random obstacles, and 3D Comb (size $110\times160\times160$) containing walls.
Planning was realized for a cubic robot of size ($10\times10\times10$).
The 3D environments are depicted in Fig.~\ref{fig:3denvs}.
As the robot can rotate and translate in 3D, the path planning leads
to a search in 6D configuration space.

We compared our planners 
(PI-RRT*, C-RRT*, PIC-RRT*, PI-BIT*, C-BIT*, PIC-BIT*)
with state-of-the-art asymptotically optimal path planners from 
the OMPL benchmark~\cite{OMPL}: Informed-RRT*, RRT*, RRTX~\cite{Otte2014RRTXRM}, RRT\#.
Despite OMPL also containing the BIT* planner, we did not use it due to its poor performance.
Therefore, we used our implementation of BIT* (with batch size 1,000) for the comparison.
We run each planner for $10^5$ iterations.
In the case of `PI-' planners, the random samples are drawn solely from the set
$\sspace_l$, as this set ensures asymptotical optimality.
In the case of planners with {`C-'} and {`{PIC-}'} prefix (i.e., drawing random
samples from $\sspace_c$ and $\sspace_{cl}$, respectively), we also generated  random samples from $\sspace_i$ with the 
the probability $10^{-5}$.
Each planner was run 100 times in each planning scenario.
The parameter $c$ was set to $5$, and goal region $\Qgoal$ was represented by a box of size $10$ units in 3D environments and of size $5$ units in 2D environments, respectively.

\input{table}

\input{table2D}

\begin{figure}
    \centering
\includegraphics[width=0.8\linewidth]{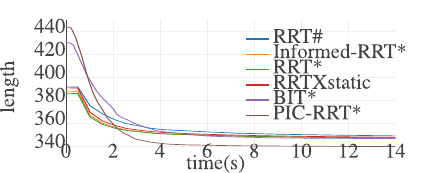}
    \caption{Convergence graph of methods in a randomly generated 3D environment.}
	\label{fig:3denvs_res}
\end{figure}

The convergence graph in Fig.~\ref{fig:3denvs_res} shows the performance of the state-of-the-art planners and our planner PIC-RRT* in the Random 3D environment.
For visibility reasons, we omitted the curves of our other planners from the graph.
From the tested planner, PIC-RRT* achieved the fastest convergence
to the optimal solution.

The results achieved in the 6D configuration space
are summarized in Tab.~\ref{tab:res_3D}.
The table shows the average path length (column 'Avg'), standard deviation (column 'Std'), and median absolute deviation (column 'Mad').
The proposed methods (the upper part of the table) found shorter paths at the given time (ten seconds of computation).

\begin{figure}
    \centering
\includegraphics[width=0.8\linewidth]{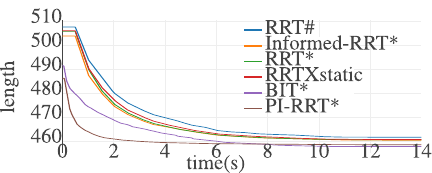}
    \caption{Convergence graph of methods in the Hard environment.}
	\label{fig:2denvs_res}
\end{figure}

The performance (the convergence towards the optimal solution)
of the planners in the Hard 2D environment is depicted in Fig.~\ref{fig:2denvs_res}.
The graph shows convergence curves for state-of-the-art planners and for our PI-RRT* planner.
We omitted our other planners (`PI-` and `PIC-') from this graph due to visibility.

The length of the paths found by the tested planners in 2D environments is summarized in Tab.~\ref{tab:res_2D}.
The table shows the path length after six seconds of runtime (after this time, most of the planners do not improve their path length).
The best average path lengths (shown in boldface in the table) were found by the proposed methods.
In the tested environments, the start and goal can be connected using topologically distinct paths.
In such cases, the first path found by the sampling-based planners may be different in each trial, and it may take a longer time to converge to the optimal one.
This is indicated by the high standard deviation, especially for Maze and Wall environments, which contain many possible ways to connect the start and goal.
Yet, our planners showed a smaller standard deviation in finding paths than the other methods.
The progress of PI-RRT* and C-RRT* is visualized in Fig.~\ref{fig:search}.

In comparison to state-of-the-art planners, the C-RRT* and C-BIT* 
planners had the worst relative performance in the Hard environment, as expected.
However, in other environments, sampling in the space $\sspace_c$ (C-RRT* and C-BIT*) enabled to find better paths than were found by other state-of-the-art planners.

In all tested 2D and 3D environments, the proposed methods outperformed 
the state-of-the-art planners: they have a faster rate of 
convergence and provide shorter paths.

\section{Conclusion}

The well-known issue of asymptotically optimal path planning using RRT* is its slow convergence towards the optimal path.
In this paper, we have proposed novel methods to approximate
the omniscient set, i.e., the subset of the configuration space
that is known to contain samples that can improve the quality of the path.
The first proposed approach uses multiple hyperellipsoids that are defined
by a subsection of the current best path.
In the second approach, we construct a convex hull of the current best path.
We describe how to sample these spaces.
The proposed methods can be integrated into any RRT*-based planner.
The experiments show the superior performance of our methods in comparison to the
state-of-the-art planners from the OMPL benchmark.

\vspace{-0.4cm}

\input{main.bbl}

\end{document}

%% file: table.tex
\begin{table}[h]
\centering
\caption{Length of path found by planners in 3D environments after ten seconds runtime.
 	\label{tab:res_3D}
 }
\vspace{-0.5em}
{
\small
\renewcommand{\arraystretch}{0.9}
 {\setlength{\tabcolsep}{4.5pt}
\begin{tabular}{lccc|ccc}
 \toprule
Environment & \multicolumn{3}{c}{Random} & \multicolumn{3}{c}{3Dcomb}\\
	Planner & Avg & Std & Mad & Avg & Std & Mad\\
\midrule
	 \textbf{C-RRT*} & 347.03  & 3.65 & 1.44 & 201.48 & 6.51 & 2.74 \\
	 \textbf{PIC-RRT*} & \textbf{340.81} & 0.81 & 0.67 & 197.84 & 5.80 & 4.90 \\
	 \textbf{PI-RRT*} & 344.15 & 13.16 & 0.76 & 198.25 & 4.09 & 5.45 \\
	 \textbf{PIC-BIT*} & 340.85 & 1.42 & 1.00 & \textbf{197.11} & 4.13 & 2.09 \\
	 \textbf{C-BIT*} & 347.10 & 1.34 & 1.27 & 197.69 & 17.00 & 2.72 \\
	 \textbf{PI-BIT*} & 341.74 & 3.31 & 0.97 & 198.07 & 4.60 & 2.85 \\
\midrule
	 \textbf{BIT*} & 348.37 & 4.74 & 2.61 & 200.55 & 1.16 & 1.02\\
	 \textbf{RRT\#} & 352.51 & 3.78 & 3.29 & 206.68 & 2.13 & 2.23 \\
	 \textbf{Informed-RRT*} & 349.45 & 4.10 & 3.11 & 204.28 & 2.95 & 3.08 \\
	 \textbf{RRT*} & 349.61 & 3.54 & 3.17 & 204.47  & 2.63 & 2.95 \\
	 \textbf{RRTXstatic} & 350.45 & 3.81 & 3.48 & 205.10 & 2.73  & 3.23 \\
\bottomrule
	\end{tabular}
 }}
 \vspace{-0.5em}
\end{table}

%% file: table2D.tex
\begin{table*}[!ht]
\centering
\caption{Length of path found by planners in 3D environments after six seconds runtime.
\label{tab:res_2D}
}\vspace{-0.5em}
{
\renewcommand{\arraystretch}{0.9}
\small
\begin{tabular}{lccc|ccc|ccc|ccc}
\toprule
Environment & \multicolumn{3}{c}{Hard} & \multicolumn{3}{c}{Maze} & \multicolumn{3}{c}{Wall} & \multicolumn{3}{c}{Comb} \\
	Planner & Avg & Std & Mad & Avg & Std & Mad & Avg & Std & Mad & Avg & Std & Mad\\
\midrule
	 \textbf{C-RRT*} & 462.27 & 11.56 & 1.58 & 534.54 & 5.09 & 3.37 & \textbf{522.53} & 27.72 & 7.39 & 552.34 & 1.51 & 1.42 \\
	 \textbf{PIC-RRT*} & 461.41 & 11.96 & 0.94 & 534.31 & 7.21 & 5.03 & 543.10 & 56.26 & 30.72 & 551.19 & 9.39 & 0.72 \\
	 \textbf{PI-RRT*} & \textbf{458.98} & 10.33 & 0.41 & 551.18 & 16.35 & 16.75 & 593.59 & 120.25 & 53.80 & 553.67 & 10.96 & 0.69 \\
	 \textbf{PIC-BIT*} & 462.43 & 12.80 & 0.95 & \textbf{533.82} & 7.92 & 3.93 & 544.20 & 54.25 & 33.37 & \textbf{550.53} & 6.29 & 0.75 \\
	 \textbf{C-BIT*} & 464.35 & 13.31 & 1.28 & 534.80 & 4.68 & 3.60 & 523.87 & 34.19 & 5.39 & 552.03 & 1.23 & 1.25 \\
	 \textbf{PI-BIT*} & 461.57 & 12.25 & 0.43 & 551.22 & 16.24 & 17.67 & 572.85 & 76.14 & 56.22 & 554.24 & 11.54 & 0.53 \\
\midrule
	 \textbf{BIT*} &461.37 &9.22 &1.69 &579.20 &13.42 &8.05 &588.77 &68.51 &79.99 &553.18 &1.44 &1.40\\
	 \textbf{RRT\#} & 464.53 & 7.64 & 2.97 & 607.58 & 28.67 & 14.41 & 705.88 & 96.73 & 96.60 & 564.32 & 1.58 & 1.30 \\
	 \textbf{Informed-RRT*} & 462.55 & 6.39 & 3.14 & 605.17 & 33.08 & 14.14 & 699.56 & 92.22 & 101.44 & 563.97 & 1.55 & 1.38 \\
	 \textbf{RRT*} & 462.69 & 6.49 & 3.01 & 602.64 & 26.78 & 12.89 & 695.13 & 90.10 & 98.33 & 563.92 & 1.55 & 1.45 \\
	 \textbf{RRTXstatic} & 463.10 & 6.77 & 2.98 & 604.89 & 28.66 & 14.54 & 699.23 & 90.04 & 99.68 & 564.05 & 1.57 & 1.40 \\
\midrule
\end{tabular}
 }
 \vspace{-2em}
\end{table*}